\def\year{2021}\relax
\documentclass[letterpaper]{article} % DO NOT CHANGE THIS
\usepackage{aaai21}  % DO NOT CHANGE THIS
\usepackage{times}  % DO NOT CHANGE THIS
\usepackage{helvet} % DO NOT CHANGE THIS
\usepackage{courier}  % DO NOT CHANGE THIS
\usepackage[hyphens]{url}  % DO NOT CHANGE THIS
\usepackage{graphicx} % DO NOT CHANGE THIS
\usepackage{graphicx}  % DO NOT CHANGE THIS
\usepackage{subfig}
\usepackage{natbib}  % DO NOT CHANGE THIS AND DO NOT ADD ANY OPTIONS TO IT
\usepackage{caption} % DO NOT CHANGE THIS AND DO NOT ADD ANY OPTIONS TO IT
\usepackage{amsmath}

\usepackage{xcolor}
\numberwithin{equation}{subsection}
\newcommand{\R}{\mathbb{R}}
\usepackage{amssymb}
\usepackage{thm-restate}
\renewcommand{\vec}[1]{\mathbf{#1}}
\newtheorem{proposition}{Property}

\usepackage[linesnumbered,ruled,vlined]{algorithm2e}

\SetCommentSty{mycommfont}

\SetKwInput{KwInput}{Input}                % Set the Input
\SetKwInput{KwOutput}{Output} 

\title{Mercer Features for Efficient Combinatorial Bayesian Optimization}
\author{Aryan Deshwal, Syrine Belakaria, Janardhan Rao Doppa\\}
\affiliations{
School of EECS, Washington State University\\
 \text{\{aryan.deshwal, syrine.belakaria, jana.doppa\}@wsu.edu}
}

\begin{document}

\maketitle

\begin{abstract}
Bayesian optimization (BO) is an efficient framework for solving black-box
optimization problems with expensive function evaluations. This paper
addresses the BO problem setting for combinatorial spaces (e.g., sequences
and graphs) that occurs naturally in science and engineering applications.
A prototypical example is molecular optimization guided by expensive
experiments. The key challenge is to balance the complexity of statistical
models and tractability of search to select combinatorial structures for
evaluation. In this paper, we propose an efficient approach referred as
{\em Mercer Features for Combinatorial Bayesian Optimization (MerCBO)}.
The key idea behind MerCBO is to provide explicit feature maps for
diffusion kernels over discrete objects by exploiting the structure of
their combinatorial graph representation. These Mercer features combined
with Thompson sampling as the acquisition function allows us to employ
tractable solvers to find next structures for evaluation. Experiments on
diverse real-world benchmarks demonstrate that MerCBO performs similarly
or better than prior methods. The source code is available at
\url{https://github.com/aryandeshwal/MerCBO}.

\end{abstract}
\section{Introduction}

The problem of black-box optimization over combinatorial spaces (e.g., sequences and graphs) using expensive evaluations arises in diverse real-world applications. A prototypical example is molecular optimization guided by wet lab experiments. Bayesian optimization (BO) is an efficient framework \cite{BO-Survey:2016} for solving such problems and has seen a lot of success for continuous spaces \cite{BO:NIPS2012,mes,MESMO,USEMO,MESMOC,MF-MESMO}. BO algorithms learn a statistical model, e.g., Gaussian Process (GP) \cite{GP-Book}, and employ it to select the sequence of inputs for evaluation guided by an acquisition function, e.g., expected improvement (EI). The key BO challenge for combinatorial spaces is to appropriately trade-off the complexity of statistical models and tractability of the search process to select structures for evaluation. Prior methods either employ simpler models that are amenable for tractable search \cite{BOCS} or perform heuristic search with complex models \cite{SMAC:TR2010,COMBO}. Additionally, these methods can only work with a limited set of acquisition functions. COMBO \cite{COMBO} is the state-of-the-art method that employs a GP-based complex statistical model using diffusion kernels and performs local search to select structures. 

In this paper, we ask the following question: {\em Can we retain the modeling strength of GP-based model using discrete diffusion kernels and still perform tractable search for selecting better structures to improve the overall BO performance?} To answer this question, we propose a novel approach referred as Mercer Features for Combinatorial Bayesian Optimization (MerCBO), pronounced as ``Merc\`{e} BO''. The main advantage of COMBO comes from the combinatorial graph representation of input space $\mathbb{X}$ that allows using diffusion kernels to define smooth functions over $\mathbb{X}$. MercBO computes a closed-form expression of this class of smooth functions in terms of explicit feature maps of diffusion kernels, which we refer as {\em Mercer features}. The key insight is to exploit the structure of the graph representation in COMBO to extract powerful low-dimensional features.  

One main advantage of Mercer features is that they allow us to leverage a large number of acquisition functions and algorithms from the continuous BO literature to improve the BO performance for combinatorial spaces. In MerCBO, we employ Thompson sampling (TS) as the acquisition function by sampling parametric functions from GP posterior via Mercer features. We show that the acquisition function optimization problem with TS is a Binary Quadratic Program (BQP). Inspired by the success of submodular relaxation in the structured prediction literature \cite{submodularization_cv}, we study a fast and scalable submdular relaxation method \cite{shinji_neurips} to solve BQPs for selecting structures. The key idea is to construct a lower bound of the BQP problem in terms of unknown relaxation parameters and iteratively optimize those parameters to improve accuracy.

In many scientific applications (e.g., biological sequence design and chemical design), we have the ability to perform a large number of parallel experiments. Therefore, there is a great need for scalable algorithms for selecting batch of inputs  for parallel evaluation to quickly uncover diverse and high-performing structures. MerCBO can be employed to meet this desiderata as Thompson sampling is embarrassingly parallel and its exploration strategy is highly helpful for diversity. Experiments on multiple real-world benchmarks including biological sequence design show that MerCBO performs significantly better than prior methods.

\vspace{1.0ex}

\noindent {\bf Key Contributions.} We develop and evaluate the MerCBO approach for BO problems over combinatorial spaces. 

\begin{itemize}
    \item Algorithm to compute efficient Mercer features for diffusion kernels to enable improved BO performance.
    
    \item Efficient submodular-relaxation approach based on Thompso
    sampling to select structures for evaluation. Demonstration of its
    scalability and effectiveness for selecting diverse batch of structures
    for parallel evaluation.

    \item Empirical evaluation using diverse real-world benchmarks and
    comparison with state-of-the-art methods. 
    
\end{itemize}

\section{Problem Setup and Related Work}

A black-box optimization problem over combinatorial spaces is specified by a 2-tuple $\left\langle \mathbb{X}, O \right\rangle$, where $\mathbb{X}$ is an input space over discrete structures (e.g., sequences, graphs) and $O$ is the black-box objective function. Without loss of generality, we assume a space of binary structures $\mathbb{X}$ = $\{0,1\}^n$. Note that we can convert any discrete space into an efficient binary representation by encoding categorical variables. Suppose each structure $\mathbf{x} \in \mathbb{X}$ be represented using $n$ binary variables $x_1, x_2,\cdots,x_n$. We are allowed to evaluate each structure $\mathbf{x} \in \mathbb{X}$ using the objective function $O: \mathbb{X} \rightarrow \R$ to observe the output $y$ = $O(\mathbf{x})$. Each function evaluation is expensive in terms of the consumed resources (e.g., computational or physical). For example, in molecular optimization application, $\mathbf{x} \in \mathbb{X}$ is a candidate molecule represented as a graph and $O(\mathbf{x} \in \mathbb{X})$ corresponds to performing a expensive wet lab experiment to measure the desired property of molecule. The overall goal is to optimize the objective $O$ by minimizing the number of evaluations.

BO methods learn a statistical model $M$ from past function evaluations and select one structure for evaluation in each iteration using an acquisition function $\alpha$ (e.g., expected improvement). To perform this selection, we need to solve the following combinatorial optimization problem:
\begin{align}
     \mathbf{x}_{next} = \mbox{arg}\max_{\mathbf{x} \in \mathbb{X}} \, \alpha(M, \mathbf{x})
     \label{eq:afo}
\end{align}

The key challenge is to appropriately trade-off the complexity of statistical model $M$ and tractability of solving the above acquisition function optimization (AFO) problem. 

\vspace{1.0ex}

\noindent {\bf Related Work.} There is limited work for BO over combinatorial spaces. Prior work addresses the above challenge by using either simple models that are amenable for tractable solvers or perform heuristic search with complex models. BOCS \cite{BOCS} employs a linear second-order model over binary variables that is amenable for semi-definite program based AFO solvers. Discrete walsh functions \cite{walshbasis_BO} are also employed to build surrogate models. SMAC \cite{SMAC,SMAC:TR2010} employs random forest as statistical model and a {\em hand-designed} local search procedure for solving AFO problems. COMBO \cite{COMBO} employs GP-based statistical model using discrete diffusion kernels that allows to model higher-order interactions between variables and uses local search with restarts as the AFO solver. COMBO was shown to perform significantly better than both SMAC and BOCS. COMEX \cite{kdd_bo} employs multi-linear polynomial as surrogate model with exponential weight updates and executes simulated annealing using the learned model to select inputs for evaluation. In a complementary direction, when AFO problems are not tractable, L2S-DISCO \cite{L2S-DISCO} exploits the relationship between the sequence of AFO problems across BO iterations via learning to improve the accuracy of search.

There is another line of work that assumes the availability of a large dataset of structures. One generic approach is to learn a continuous representation using this dataset and to perform BO in the latent space \cite{reduction_continuous}. One potential drawback of this reduction method is that it can generate invalid structures  \cite{janz2017learning,grammar_vae}. There are also methods that employ this big data to learn Bayesian neural networks as statistical models \cite{parallel_ts_lobato}. Unfortunately, this assumption is violated in %many real-world
applications where small data is the norm.

\section{MerCBO Algorithm}

\begin{algorithm}
\footnotesize
  \KwInput{Input space $\mathbb{X}$, Black-box objective $O$, Order of Mercer features $\textsc{Max}$, Query budget $B$}
  \KwOutput{Best input structure found $\mathbf{x}_{best}$}
 \DontPrintSemicolon
\SetKwProg{myproc}{}{}{}
\myproc{Mercer\_Features(input structure $\mathbf{x}$, order $\textsc{Max}$)}{
    \nl Initialize  empty list ${\phi}$ = []\;
    \For{i=1 to  $\textsc{Max}$}{
    Compute the features for the $i$th order{\bf:} $\{\sqrt{e^{-\beta \lambda_i}} \cdot {-1}^{<r_i, \mathbf{x}>}\}$\ and append to ${\phi}$\
    }
    \KwRet{Mercer features $\phi$} \
    }{}{}
  \setcounter{AlgoLine}{0}
   Initialize a small-sized training set $\textsc{Train}$ with Mercer features computed for input structures \\
   \While{Query budget does not exceed $B$}
   {
        Learn Gaussian Process model $M$ using $\textsc{Train}$ \;
        {\color{blue}Construct Thompson sampling acquisition function by sampling from a parametric approximation of the GP posterior\;
        Find $\mathbf{x}_{next}$ by optimizing Thompson sampling based acquisition function over model $M$\;
       \SetAlgoLined \Begin{
        {\bf Step 1:} Construct submodular relaxation based lower bound of AFO with parameters\;
        \indent \While{convergence or maximum iterations}{
            {\bf Step 2:} Solve the relaxed problem using graph cut algorithms \;
            {\bf Step 3:} Update the relaxation parameters to obtain a tighter bound \;
        }}
        }
        Evaluate $O$ at $\mathbf{x}_{next}$ and add to $\textsc{Train}$: $\mathbf{x}_{next}$, Mercer\_Features($\mathbf{x}_{next}$, $\textsc{Max}$), and $O(\mathbf{x}_{next})$\;
   }
%     \tcc{Now this is a While loop}

\Return{Best input $\mathbf{x}_{best}$ and its objective value $O(\mathbf{x}_{best})$}
\caption{MerCBO Algorithm}
\label{algo:mercbo}
\end{algorithm}

\noindent We first provide an overview of MerCBO and its advantages. 

\vspace{0.8ex}

\noindent {\bf Overview of MerCBO algorithm.} MerCBO is an instantiation of the generic BO framework. 1) Gaussian Process with diffusion kernels is the surrogate statistical model. 
2) Thompson sampling is used as the acquisition function $\alpha$. The proposed Mercer features are used to sample from a parametric approximation of the GP posterior to construct the Thompson sampling objective. 3) Acquisition function optimization problem is shown to be a Binary Quadratic Program which is solved using an efficient and scalable submodular-relaxation method. The key idea is to construct a lower bound of the AFO problem in terms of some unknown relaxation parameters and iteratively optimize those parameters to obtain a tighter bound. Algorithm~\ref{algo:mercbo} shows the complete pseduo-code of MerCBO. We use a small set of input structures and their function evaluations (denoted \textsc{Train}) to bootstrap the surrogate model. In each iteration, we select the next structure for evaluation $\mathbf{x}_{next}$ by solving the AFO problem; add $\mathbf{x}_{next}$, its function evaluation $O(x_{next})$, and mercer features of $\mathbf{x}_{next}$ to \textsc{Train}. We repeat these sequence of steps until the query budget is exhausted and then return the best found structure $\mathbf{x}_{best}$ as the output.

\vspace{0.8ex}

\noindent {\bf Advantages of MerCBO} over COMBO and SMAC. {\bf 1)} Mercer features allow us to leverage a large number of acquisition functions from the continuous BO literature including Thompson sampling (TS), PES \cite{pes} and MES \cite{mes} to improve the BO performance for combinatorial spaces. {\bf 2)} Retains the modeling strength of complex GP-based model using diffusion kernels and still allows a tractable and scalable AFO procedure with Thompson sampling. {\bf 3)} Mercer features combined with TS has many desiderata required for several scientific applications: {\em diversity} in explored structures and selection of {\em large batch} of structures for parallel evaluation. Indeed, our experiments on biological sequence design demonstrate the effectiveness of TS (embarrassingly parallel) over COMBO with EI.

\subsection{Preliminaries}

\noindent {\bf Graph Laplacian.} Given a graph $G$ = $(V, E)$, its Laplacian matrix $L(G)$ is defined as $D-A$, where $D$ is the degree matrix and $A$ is the adjacency matrix of $G$.

\vspace{0.8ex}

\noindent {\bf Graph Cartesian Product.} Given two graphs $G_1$ = $(V_1, E_1)$ and $G_2$ = $(V_2, E_2)$, their graph Cartesian product $(G_1 \square G_2)$ is another graph with vertex set $V(G_1 \square G_2)$ = $V_1 \times V_2$ consisting of the set Cartesian product of $V_1$ and $V_2$. Two vertices $(v_1, u_1)$ and $(v_2, u_2)$  of $G_1 \square G_2$ (where $\{v_1, v_2\} \in V_1$ and $\{u_1, u_2\} \in V_2$) are connected if either, $v_1$ = $v_2$ and $(u_1, u_2) \in E_2$, or  $u_1$ = $u_2$ and $v_1, v_2 \in E_2$. 

\vspace{0.8ex}

\noindent {\bf Combinatorial Graph Representation of Discrete Space.} Recall that we need a graph representation of the discrete space $\mathbb{X}$ to employ diffusion kernels for learning GP models. We consider the combinatorial graph representation (say $G$=$(V, E)$) proposed in a recent work \cite{COMBO}. There is one vertex for each candidate assignment of $n$ discrete variables $x_1, x_2,\cdots,x_n$. There is an edge between two vertices if and only if the Hamming distance between the corresponding binary structures is one.
This graph representation was shown to be effective in building GP models for BO over discrete spaces \cite{COMBO}.

\subsection{Efficient Mercer features for Diffusion Kernel} 

We are interested in computing explicit feature maps for diffusion kernel over discrete spaces \cite{diffusion_kernel_original,COMBO}, which is defined using the above combinatorial graph representation as follows:
\begin{align}
K(V, V) &= U \exp(-\beta \Lambda) U^T 
\end{align}
where $U$ = $[u_0, u_1, \cdots, u_{2^n-1}]$ is the eigenvector and $\Lambda$ = $[\lambda_0, \lambda_1, \cdots, \lambda_{2^n-1}]$ is the eigenvalue matrix of the graph Laplacian $L(G)$ and $\beta$ is a hyper-parameter. For any two given structures $\mathbf{x}_1, \mathbf{x}_2 \in \{0,1\}^n$, the kernel definition is:
\begin{align}
    K(\mathbf{x}_1, \mathbf{x}_2) = \sum_{i=0}^{2^n-1} e^{-\beta \lambda_i} u_i([\mathbf{x}_1]) u_i([\mathbf{x}_2]) \label{main_eqn}
\end{align}
where $u_i([\mathbf{x}_1])$ denotes the value of the $i$th eigenvector indexed at the integer value represented by $\mathbf{x}_1$ in base-2 number system. From the above equation, we can see that the eigenspace of the graph Laplacian $L(G)$ is the central object of study for diffusion kernels. The key insight behind our proposed approach is to exploit the structure of the combinatorial graph $G$ and its Laplacian $L(G)$ for computing its eigenspace in a {\em closed-form}. 

Since $G$ is an exponential-sized graph with $2^n$ vertices, computing the eigenspace of $L(G)$ seems intractable at first sight. However, $G$ has special structure: $G$ has an equivalent representation in terms of the graph Cartesian product of $n$ sub-graphs $G_1, G_2,\cdots,G_n$:
\begin{align}
    G &= (((G_1 \square G_2)\square G_3 \cdots)\cdots )\square G_n) \label{cg_gc}
\end{align}
where each sub-graph $G_i$ represents the $i$th binary input variable and is defined as a graph with two vertices (labeled 0 and 1) and an edge between them. Therefore, $L(G)$ is equivalently given by:
\begin{align}
    L(G) &= L((((G_1 \square G_2)\square G_3 \cdots)\cdots )\square G_n)) \\
    L(G) &=  L(G_1) \oplus L(G_2) \oplus L(G_3) \cdots \oplus L(G_n) \label{laplacian_cartesian_product}
\end{align}
where Equation \ref{laplacian_cartesian_product} is due to distributive property of the Laplacian operator over graph Cartesian product via Kronecker sum operator ($\oplus$) \cite{product_graphs_handbook} and associative property of the $\oplus$ operator.

\begin{proposition} \label{gc_eigenspace_property} \cite{product_graphs_handbook}
Given two graphs $G_1$ and $G_2$ with the eigenspace of their Laplacians being $\{\Lambda^1, U^1\}$ and $\{\Lambda^2, U^2\}$  respectively, the eigenspace of $L(G_1 \square G_2)$ is given by $\{\Lambda^1  {\bowtie} \Lambda^2, U^1 \otimes U^2\}$ where $\Lambda^1 {\bowtie} \Lambda^2 = \{\lambda_i^1 + \lambda_j^2 : \lambda_i^1 \in \Lambda^1, \lambda_i^2 \in \Lambda^2\}$ and $\otimes$ is the Kronecker product operation.
\end{proposition}

Property \ref{gc_eigenspace_property} gives a recursive procedure to compute the eigenspace of $L(G)$ based on its decomposition in Equation \ref{laplacian_cartesian_product} provided the eigenspace of each of its constituent $L(G_i)$ is easily computable. Fortunately, in our special case, where each $G_i$ is a simple graph of two vertices (labeled 0 and 1) with an edge between them, it has two eigenvalues $\{0, 2\}$ with corresponding eigenvectors $[1, 1]$ and $[1, -1]$ respectively. The eigenvector matrix $\begin{bmatrix}
 1 & 1 \\ 1 & -1  \end{bmatrix}$ is called as a Hadamard matrix of order 2 $(2^1$, where $1$ in the exponent is the number of input variables) \cite{sgt_hypercube, bamk_paper}. Applying the $\bowtie$ and $\otimes$ operation recursively for $n$ inputs (represented by $\{G_i: i \in \{1, 2, \cdots, n\}\}$) as described in property \ref{gc_eigenspace_property}, it can be seen that the eigenspace of L(G) has an explicit form given as:
 
\begin{enumerate}
\item $L(G)$ has $n$ distinct eigenvalues $\{0, 2, 4, \cdots, 2n\}$ where $j$th eigenvalue occurs with multiplicity $\binom{n}{j}, j \in \{0, 1, 2, \cdots n\}$.
\item The eigenvectors of $L(G)$ are given by the columns of Hadamard matrix of order $2^n$. 
\end{enumerate}

\noindent Hadamard matrix of order $2^n$ is equivalently defined as:
\begin{align}
    H_{ij} = (-1)^{<r_i, r_j>}
\end{align}
where $r_i$ is the $n$-bit representation of the integer $i$ in base-2 system. Using this definition of Hadamard matrix, each entry of the eigenvectors of $L(G)$ can be computed in a closed form.
Therefore, from Equation \ref{main_eqn}, the kernel value for any pair of structures $\mathbf{x}_1$ and $\mathbf{x}_2$, $K(\mathbf{x}_1, \mathbf{x}_2)$, can be written in terms of an equivalent sum over binary vectors:
\begin{align}
K(\mathbf{x}_1, \mathbf{x}_2) &= \sum_{i=0}^{2^n-1} e^{-\beta \lambda_i} \cdot {-1}^{<r_i, \mathbf{x}_1 + \mathbf{x}_2>} \label{first_eqn_with_r}
\end{align}
where $r_i$ is the base-2 representation of integer $i$ ranging from 0 to $2^n$-1. %Should we keep $0$-based indexing throughout?
We rearrange the RHS of Equation \ref{first_eqn_with_r} to  delineate the dependency on each input in the pair.
\begin{align}
K(\mathbf{x}_1, \mathbf{x}_2) &= \sum_{i=0}^{2^n-1} e^{-\beta \lambda_i} \cdot {-1}^{<r_i, \mathbf{x}_1>} \cdot{-1}^{<r_i, \mathbf{x}_2>} \\
K(\mathbf{x}_1, \mathbf{x}_2) &= <\phi(\mathbf{x}_1), \phi(\mathbf{x}_2)>
\label{main_feature}
\end{align}
where $\phi(\mathbf{x})$ corresponds to the proposed (explicit) {\em Mercer feature maps} of any input structure $\mathbf{x}$ and is given as follows:
\begin{align}
    \phi(\mathbf{x}) = \{i \in [0, 2^n - 1]: \sqrt{e^{-\beta \lambda_i}} \cdot {-1}^{<r_i, \mathbf{x}>}\} \label{full_feature_maps}
\end{align}

As mentioned earlier, the eigenvalues of $L(G)$ are explicitly given by the set $\{2j: j \in \{0, 1, 2, \cdots n\}\}$, where $j$th eigenvalue occurs with multiplicity $\binom{n}{j}$. Based on this observation, an elegant way of interpreting the feature maps given in \ref{full_feature_maps} is based on the number of 1s in the $r_i$ vector (binary expansion of integer $i$). There are exactly $\binom{n}{j}$ $r$-vectors with $j$ bits set to 1. Hence, we refer to $j$ as the {\em order of the Mercer feature maps}. The order variable controls the trade-off between the computational cost and approximation quality of the feature map. We found that the second-order feature maps\footnote{Slight abuse of notation. Second-order, from here-on, means concatenation of both first and second-order features.} maintain the right balance as they can be computed efficiently and also allows to perform a tractable and scalable search for acquisition function optimization as described in the next section.  Moreover, choosing second order is also prudent from the viewpoint of the definition of diffusion kernels itself, which requires suppressing higher frequency elements of the eigenspace \cite{COMBO} to define a smooth function over discrete spaces.

\vspace{1.0ex}

\subsection{Tractable Acquisition Function Optimization} \label{sec:tractable_afo}

In this section, we describe a tractable and scalable acquisition function optimization  procedure using the proposed Mercer features and Thompson sampling.  
Thompson sampling \cite{thompson_sampling,parallel_ts_lobato,kandasamy2017asynchronous} selects the next point for evaluation by maximizing a sampled function from the GP posterior. We approximate the non-parametric GP model using a parametric Bayesian linear model $f(\mathbf{X})$ = $\theta^T \phi(\mathbf{x})$ using our proposed Mercer features $\phi(\mathbf{x})$. Given a Gaussian prior, i.e., $\theta \sim \mathcal{N}(0, I)$, the posterior distribution over $\theta$ is also a Gaussian with the following mean and covariance:
\begin{align}
    \vec{\mu} &= \vec{(\Phi^T\Phi + \sigma^2I)^{-1} \Phi^T y}\\
    \vec{\Sigma} &= \vec{(\Phi^T\Phi + \sigma^2I)^{-1}}\sigma^2 \label{dist_params}
\end{align}
with $\vec{\Phi}$ is the feature matrix with $i$th row corresponding to $\phi(\mathbf{x}_i)$ and $\vec{y}$ is the output vector with $y_i$ $\sim$ $\mathcal{N}(f(x_i), \sigma^2)$. 

\vspace{1.0ex}

\noindent {\bf Acquisition function optimization problem.} We sample $\theta^*$ from the posterior parametrized by $\vec{\mu}$ and $\vec{\Sigma}$ defined in Equation \ref{dist_params} and minimize the objective $f(\mathbf{x})$ = $\theta^*\phi(\mathbf{x})$ with respect to $\mathbf{x} \in \{0, 1\}^n$. Suppose $\mathbf{x}$=$(x_1,x_2,\cdots,x_n)$ is a candidate structure with values assigned to $n$ binary variables. We now show how this AFO problem is an instance of Binary quadratic programming (BQP) problem by using second-order feature maps from \ref{full_feature_maps}. 
\begin{align}
        \phi(\mathbf{x}) = \left\{0 \leq i \leq {\binom{n}{2}}: \sqrt{e^{-\beta \lambda_i}} \cdot {-1}^{<r_i, \mathbf{x}>}\right\}
\end{align}

The second-order feature maps are composed of two major parts: features constructed by order-1 $r$-vectors and features constructed by order-2 $r$-vectors. For order-1 $r$-vectors, the set $\{0 \leq i \leq n: {-1}^{<r_i, \mathbf{x}>}\}$ is equivalent to $\{0 \leq i \leq n :{-1}^{x_i}\}$. Similarly, the order-2 feature maps can also be written as $\{1 \leq i \leq n, i+1 \leq j \leq n:{-1}^{(x_i + x_j)}\}$. Therefore, ignoring the constant corresponding to the $0$th index of $\theta^*$, the AFO problem $\min_{\mathbf{x} \in \{0, 1\}^n} \theta^* \phi(\mathbf{x})$ becomes:
\begin{align}
\begin{split}    
\min_{\mathbf{x} \in \{0, 1\}^n} &  \sum_{i=1}^n \theta_i^* \sqrt{e^{-\beta \lambda_i}}\cdot {-1}^{x_i} \\
& + \sum_{i=1}^n \sum_{j=i+1}^n \theta_{ij}^* \sqrt{e^{-\beta \lambda_{n\cdot i+j}}}  \cdot {-1}^{(x_i + x_j)} \label{main_afo}
\end{split}
\end{align}

By replacing the ${-1}^{x_i}$ terms in \ref{main_afo} with an equivalent term $(1-2x_i)$, we get:
\begin{align}
\begin{split}
        \min_{\mathbf{x} \in \{0, 1\}^n} & \sum_{i=1}^n \theta_i^* \sqrt{e^{-\beta \lambda_i}} (1-2x_i)\\
        &+ \sum_{i=1}^n \sum_{j=i+1}^n \theta_{ij}^* \sqrt{e^{-\beta \lambda_{n\cdot i+j}}}  (1-2x_i)(1-2x_j)  \\
\end{split}
\end{align}
Rearranging and combining terms with the same degree, we get the following final expression for AFO which is clearly a Binary quadratic programming (BQP) problem:
\begin{align}
    \min_{\mathbf{x} \in \{0, 1\}^n} \vec{b}^T \mathbf{x} + x^T \vec{A} \mathbf{x} \label{main_bqp_objective}
\end{align}
where $\vec{b}$ and $\vec{A}$ are defined as given below:

\begin{align}
    b_i &= -2 \left(\theta_i^* \sqrt{e^{-\beta \lambda_i}} + \sum_{j=1}^n \theta_{ij}^* \sqrt{e^{-\beta \lambda_{n\cdot i+j}}}\right), {1 \leq i \leq n} \\
    A_{ij} &= \delta_{ij} \cdot 4 \left( \theta_{ij}^* \sqrt{e^{-\beta \lambda_{n\cdot i+j}}}\right), 1 \leq i,j \leq n
\end{align}
where $\delta_{ij}$ = $1$ if $j > i$ and $0$ otherwise.

\vspace{1.0ex}

\noindent {\bf Efficient submodular-relaxation solver.} BQP is a well-studied problem in multiple areas including computer vision \cite{submodularization_cv}.
Motivated by the prior success of submodular-relaxation methods in the structured prediction area \cite{submodular_cv_1}, we study a fast and scalable approach for solving AFO problems based on recent advances in submodular-relaxation \cite{shinji_neurips}. The key idea is to construct an efficient relaxed problem with some unknown parameters and optimize those parameters iteratively to improve the accuracy of solutions. We provide a high-level sketch of the overall algorithm \cite{PSR} below.

The objective in \ref{main_bqp_objective} is called as {\em submodular} if $A_{ij} \leq 0 \quad \forall i, j$. Submodular functions can be exactly minimized by fast strongly polynomial-time graph-cut algorithms \cite{fujishige_book}. However, in general, the objective might not be submodular. Therefore, a submodular relaxation to the objective in Equation \ref{main_bqp_objective} \cite{shinji_neurips} is constructed by lower bounding the positive terms $A^+$ of $A$:
\begin{align}
\begin{split}
  {\mathbf{x}}^T (\vec{A}^+ \circ \Gamma) \vec{1} + \vec{1}^T (\vec{A}^+ \circ \Gamma) {\mathbf{x}} - \vec{1}^T (\vec{A}^+ \circ \Gamma) \vec{1}  \\
\leq  \mathbf{x}^T \vec{A}^+ \mathbf{x}
\end{split}
\end{align}
where $\mathbf{A}$ = $\mathbf{A}^+ + \mathbf{A}^-$,
\begin{align}
\mathbf{A}^+ &= {A_{ij} \text{ if}  A_{ij} \geq 0 \text{ and } 0 \text{ otherwise}} \\
\mathbf{A}^- &= {A_{ij} \text{ if}  A_{ij} \leq 0 \text{ and } 0 \text{ otherwise}} 
\end{align}
and $\Gamma$ stands for unknown relaxation parameters. The accuracy of this relaxed problem critically depends on $\Gamma$. Therefore, we perform optimization over $\Gamma$ parameters after initializing them by repeating the following two steps.

\begin{enumerate}
    \item Solve submodular relaxation lower bound of the BQP objective in \ref{main_bqp_objective} using minimum graph-cut algorithms.
    \item Update relaxation parameters $\Gamma$ via gradient descent
\end{enumerate}
This submodular-relaxation based AFO solver scales gracefully with the increasing input dimension $n$ because of the strongly polynomial complexity $O(n^3)$ of minimum graph cut algorithms \cite{network_flows}. 
\section{Experiments and Results}

%In this section, we empirically evaluate MerCBO, compare with state-of-the-art methods, and discuss the results.

\vspace{0.8ex}

\noindent {\bf Experimental setting.} The source code for state-of-the-art baseline methods was taken from their github links: COMBO (\url{https://github.com/QUVA-Lab/COMBO}),  BOCS (\url{https://github.com/baptistar/BOCS}), and SMAC (\url{https://github.com/automl/SMAC3}). For a fair comparison, the priors for all GP hyper-parameters and their posterior computation were kept the same for both COMBO and MerCBO. COMBO employs a separate hyper-parameter $\beta_i$ for each dimension to enforce sparsity, which is important in certain applications. For MerCBO also, we include sparsity in sampling $\theta$ for Thompson Sampling for all benchmarks other than Ising: $\mathcal{N}\left(\vec{(\Phi^T\Phi + \sigma^2 \Upsilon^{-1})^{-1} \Phi^T y},\vec{(\Phi^T\Phi + \sigma^2 \Upsilon^{-1})^{-1}}\sigma^2\right)$, where $\Upsilon$ = $\text{diag}(\beta_1, \beta_2, \cdots, \beta_1 \beta_2, \cdots, \beta_{n-1} \beta_{n})$. It should be noted that we are not introducing any more hyper-parameters than COMBO, but use them in a {\em strong heirarchical} sense \cite{bien2strong_heirarchy_bayesian_2}.

We ran five %outer 
iterations of submodular relaxation approach for solving AFO %described in Section \ref{sec:tractable_afo} 
problems and observed convergence. We ran each method on all the benchmarks for 25 random runs and report mean and two times the standard error for results.

\subsection{Sequential Design Optimization Benchmarks}
\begin{figure*}[h!]
\centering
\subfloat[Subfigure 2 list of figures text][]{
\includegraphics[width=0.25\textwidth]{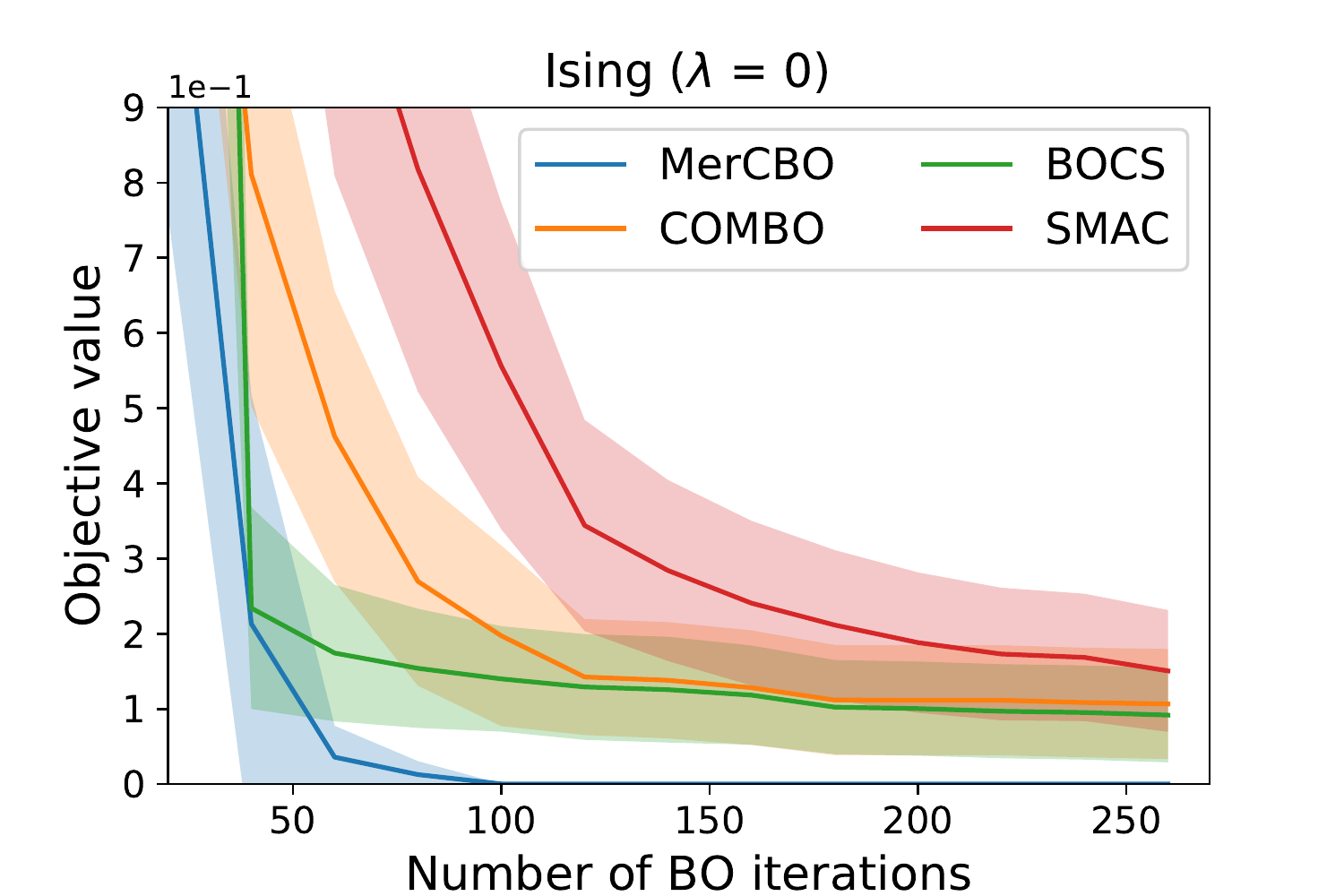}
\label{fig:ising_0}}
\subfloat[Subfigure 1 list of figures text][]{
\includegraphics[width=0.25\textwidth]{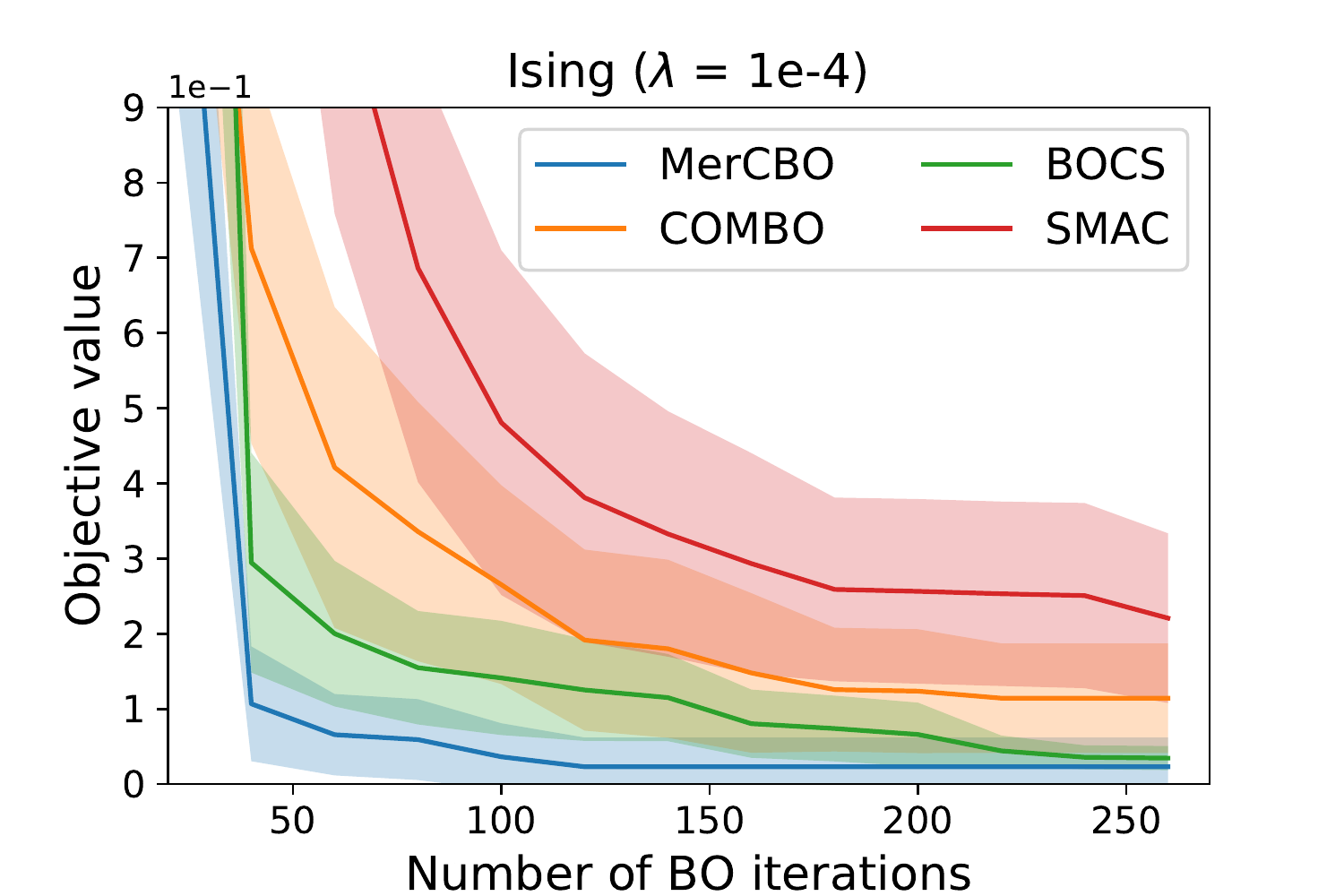}
\label{fig:ising_1e_4}}
\subfloat[Subfigure 2 list of figures text][]{
\includegraphics[width=0.25\textwidth]{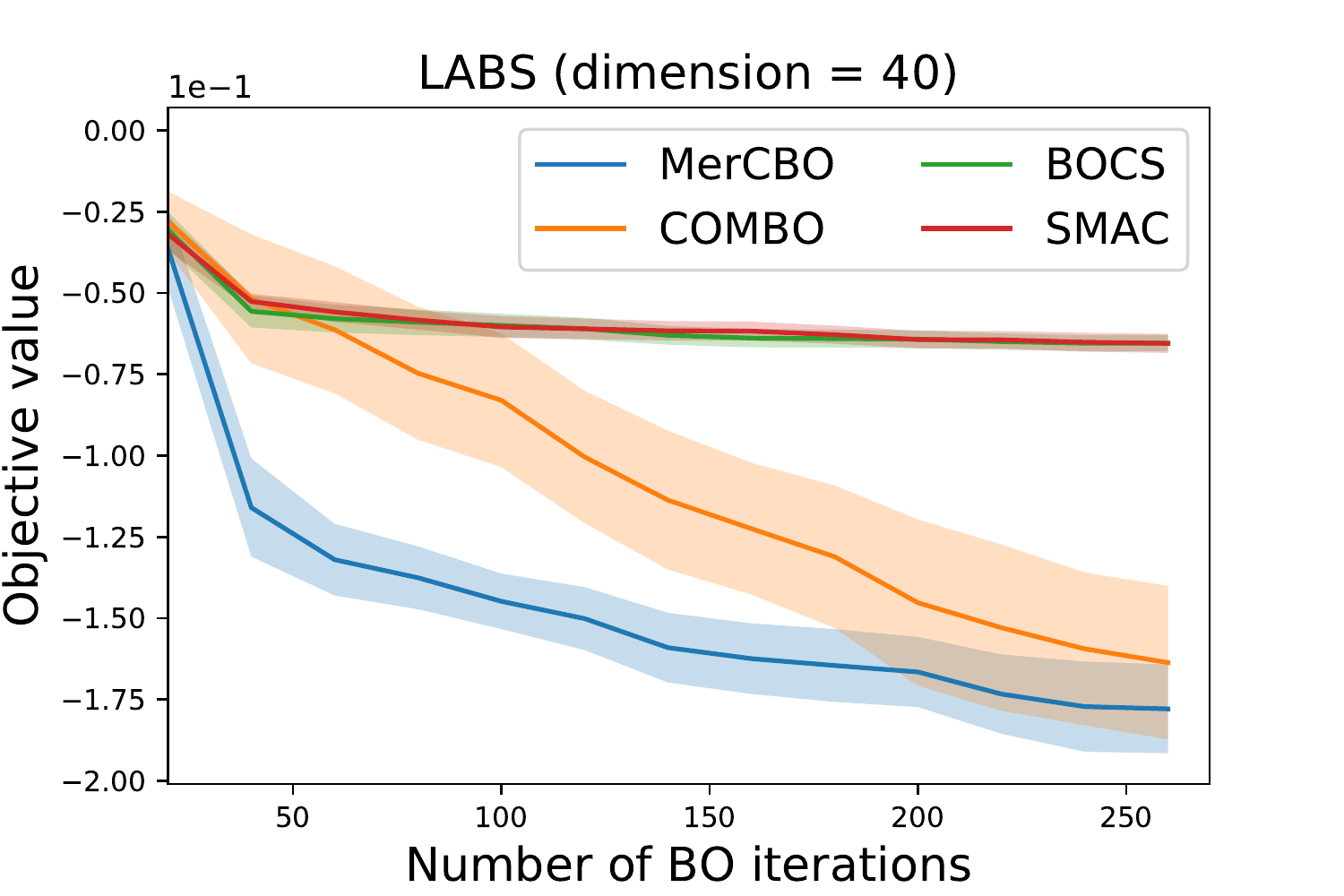}
\label{fig:labs_40}}
\subfloat[Subfigure 2 list of figures text][]{
\includegraphics[width=0.25\textwidth]{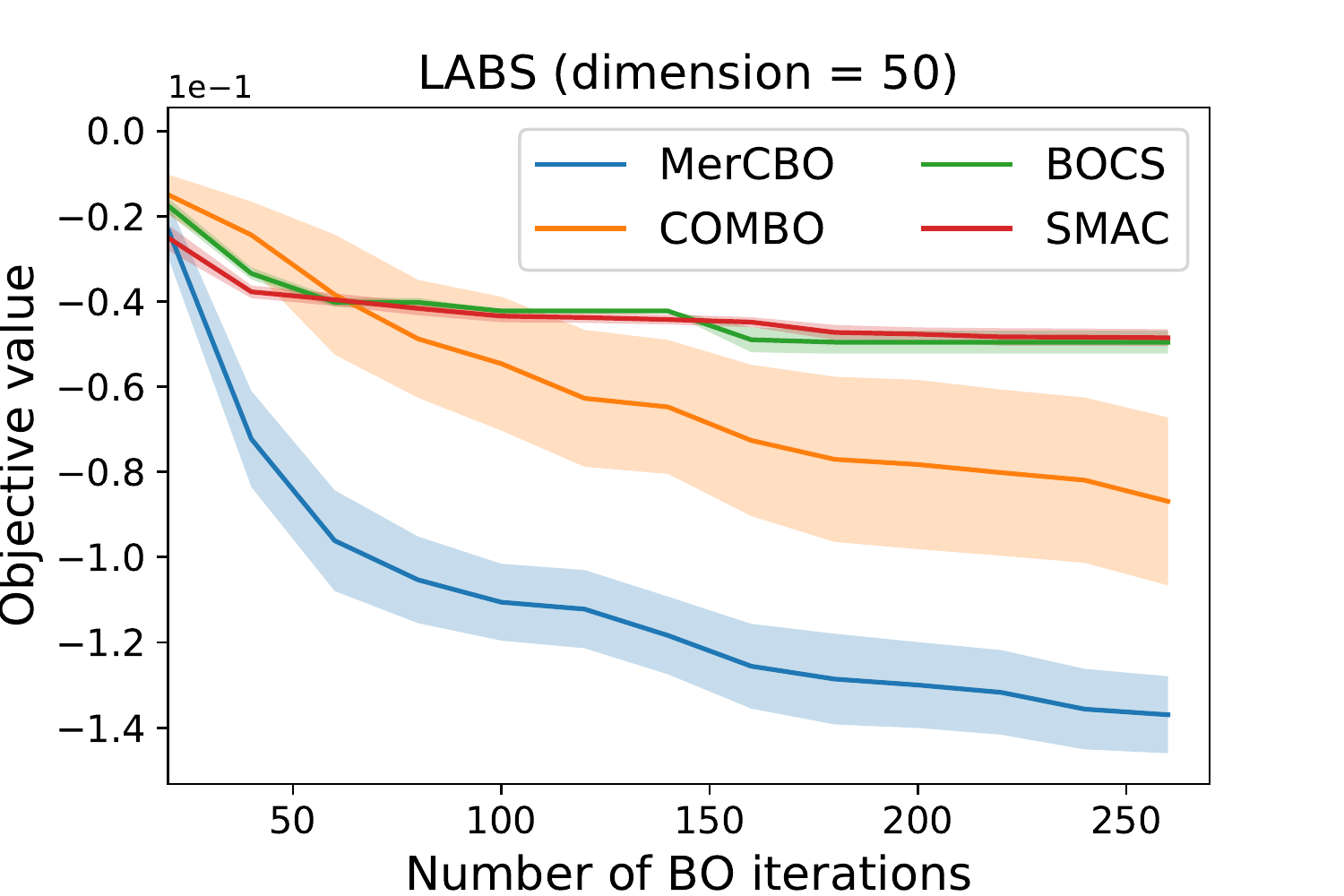}
\label{fig:labs_50}}
\caption{Results for Ising and LABS domains.} 
\label{fig:ising_labs}
\end{figure*}

We employed four diverse benchmarks for sequential design: {\em function evaluations are performed sequentially}. %one after another.

\vspace{0.8ex}

\noindent {\bf Ising sparsification.} The probability distribution $p(z)$ \cite{BOCS,COMBO} defined by Zero-field Ising model $I_p$ is parametrized by a symmetric interaction matrix  $J^p$ whose support is represented as a graph $G^p$. The goal in this problem is to approximate $p(z)$ with another distribution $q(z)$ such that the number of edges in $G^q$ are minimized. The objective function is defined as:
\begin{align*}
    \min_{\mathbf{x} \in \{0, 1\}^n} D_{KL} (p||q) + \lambda \|\mathbf{x}\|_1
\end{align*}
where $D_{KL}$ is the KL-divergence between $p$ and $q$, and $\lambda$ is a regularization parameter. The results for this $24$-dimensional domain are shown in Figure \ref{fig:ising_0} and \ref{fig:ising_1e_4}. In COMBO \cite{COMBO}, BOCS was shown to perform better than COMBO on this domain. However, MerCBO shows the best performance among all methods signifying that the proposed approach augments the performance of GP surrogate model. 

\begin{figure}[h!]
\centering
\subfloat[Subfigure 2 list of figures text][]{
\includegraphics[width=0.25\textwidth]{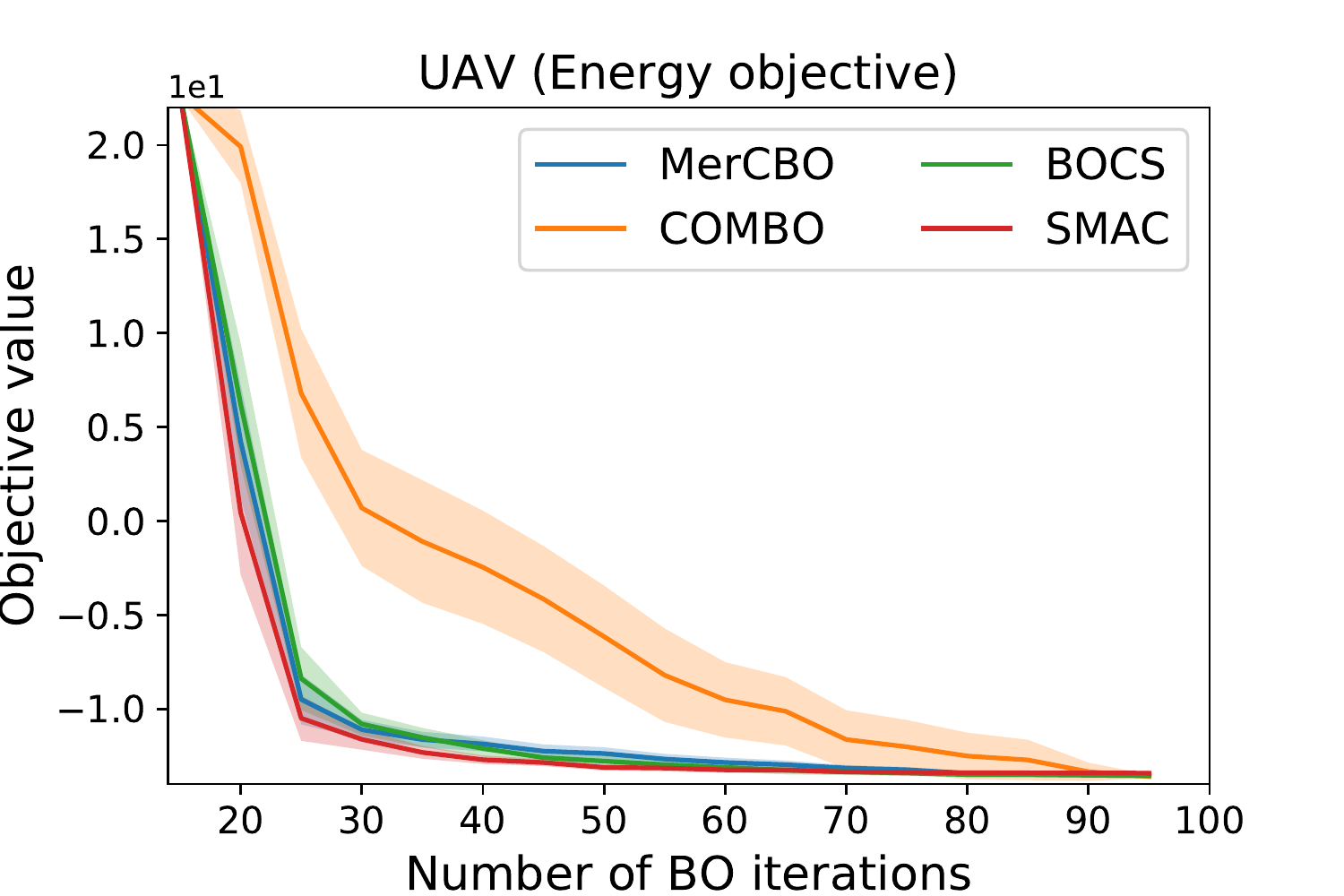}
\label{fig:uav_energy}}
\subfloat[Subfigure 1 list of figures text][]{
\includegraphics[width=0.25\textwidth]{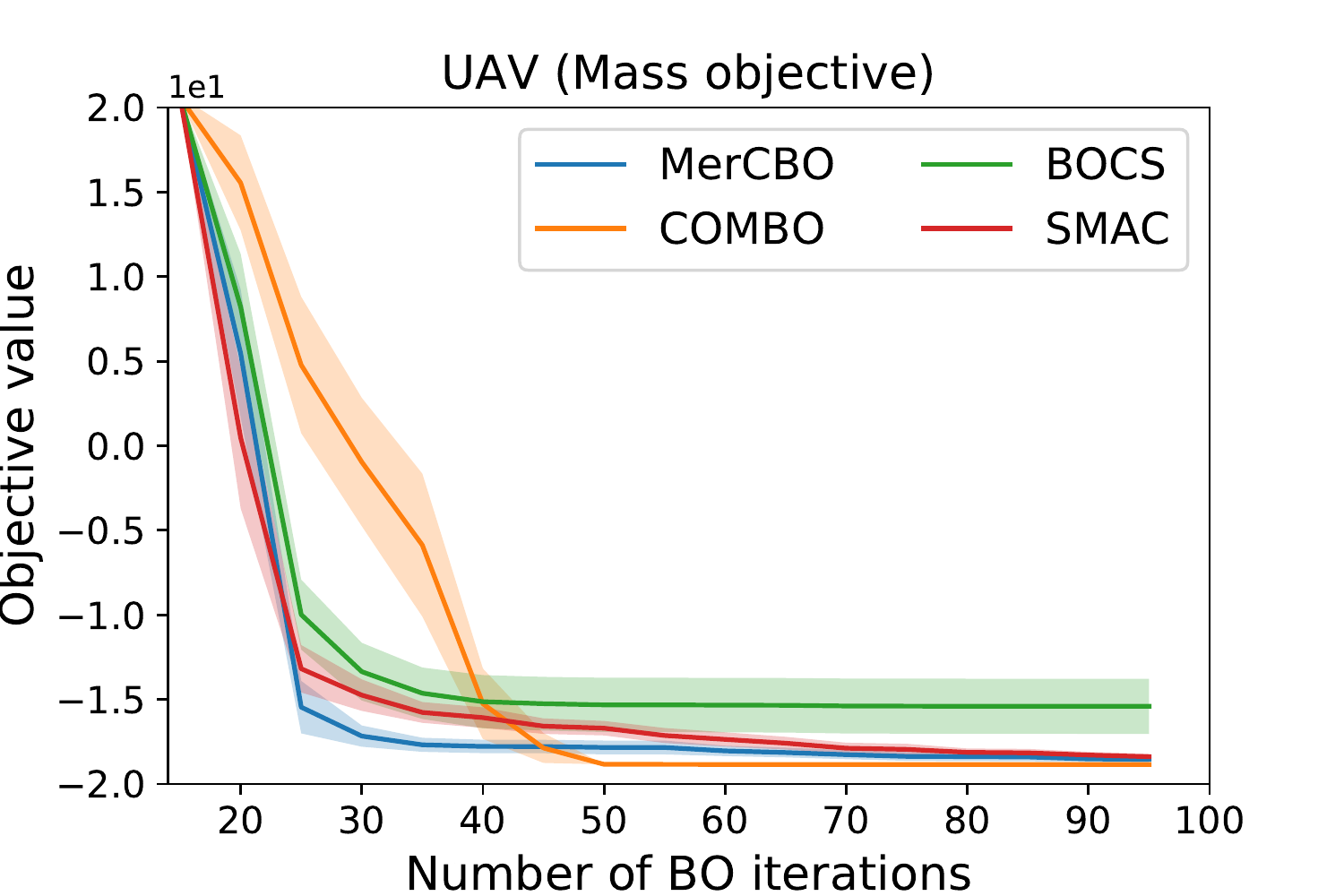}
\label{fig:uav_mass}}
\caption{Results for power system design of UAVs.} 
\label{fig:uav}
\end{figure}

\vspace{0.8ex}

\noindent {\bf Low auto-correlation binary sequences (LABS).} This problem has diverse applications in multiple fields \cite{LABS_statistical_physics,LABS_main} including communications where it is used in  high-precision interplanetary radar measurements \cite{LABS_communication}. The goal is to find a binary sequence $\{1, -1\}$ of length $n$ that maximizes the  {\em Merit factor (MF)} defined as follows:
\begin{align*}
\max_{\mathbf{x} \in \{1, -1\}^n} \text{MF($\mathbf{x}$)} &= \frac{n^2}{E(\mathbf{x})} \hspace{1mm}, \\
E(\mathbf{x}) &= \sum_{k=1}^{n-1} \left(\sum_{i=1}^{n-k} x_i x_{i+k}\right)^2
\end{align*}
where $E(\mathbf{x})$ is the energy of the sequence. This domain allows us to evaluate all methods on large input dimensions. Results with 40 and 50 dimensions are shown in Figure \ref{fig:labs_40} and \ref{fig:labs_50} respectively.  MerCBO shows best performance among all methods showing its effectiveness. COMBO shows poor convergence and matches MerCBO only on 40 dimensional case at the end of allocated budget. SMAC and BOCS show poor performance resulting from their inability to model the underlying structure properly.

\begin{figure*}[h!]
\centering
\subfloat[Subfigure 2 list of figures text][]{
\includegraphics[width=0.33\textwidth]{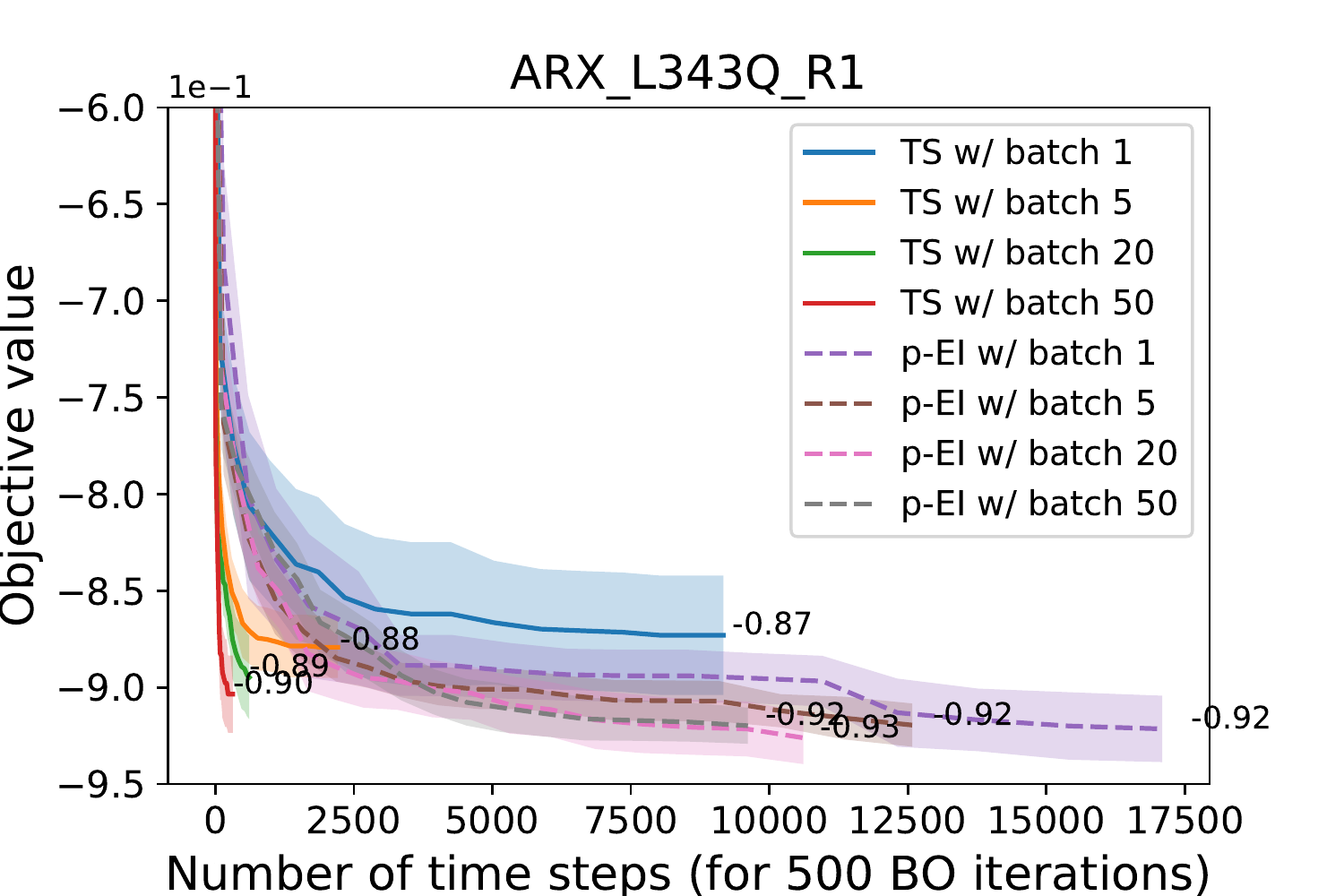}
\label{fig:bio_seq_time}}
\subfloat[Subfigure 1 list of figures text][]{
\includegraphics[width=0.33\textwidth]{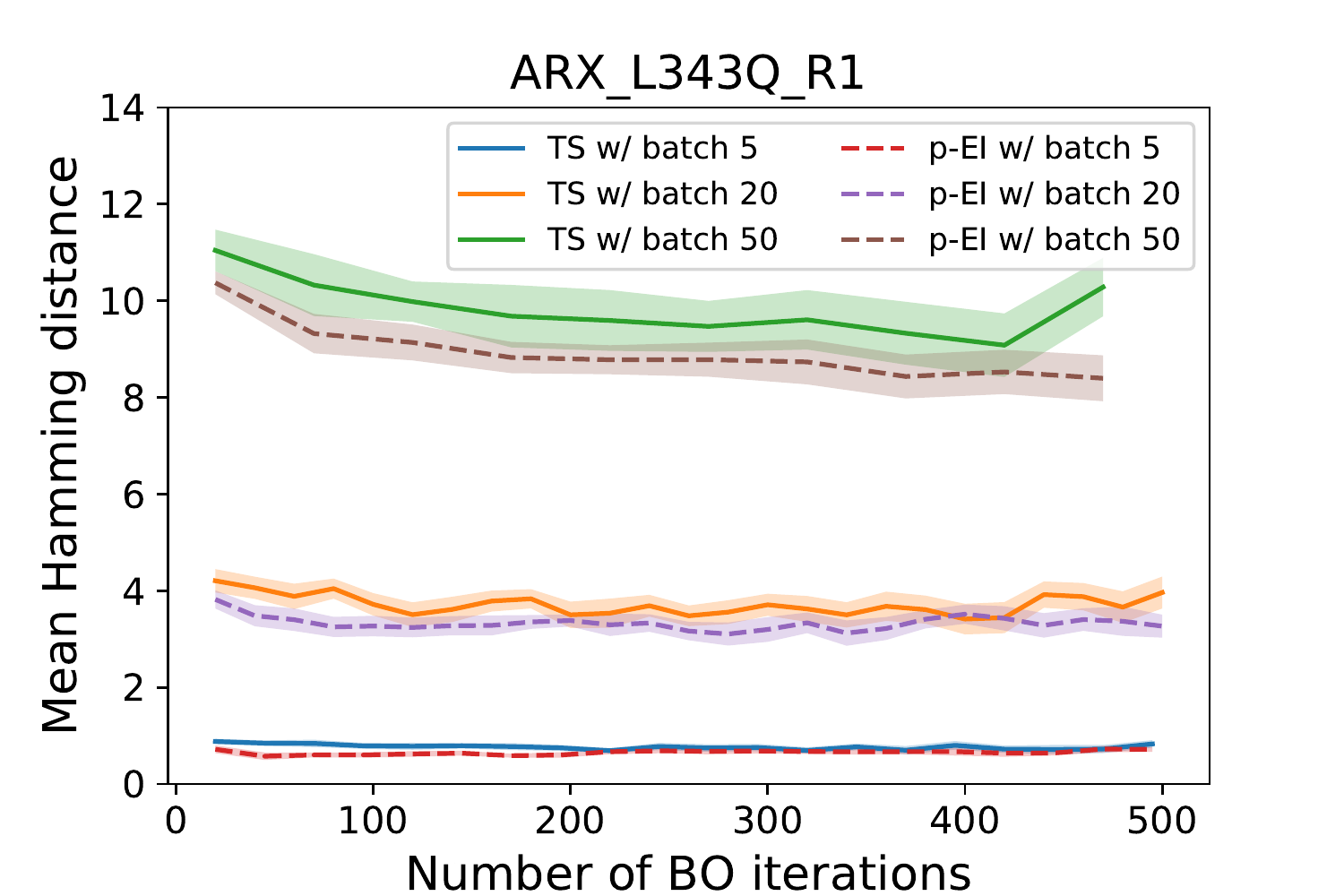}
\label{fig:bio_seq_diversity}}
\subfloat[Subfigure 2 list of figures text][]{
\includegraphics[width=0.33\textwidth]{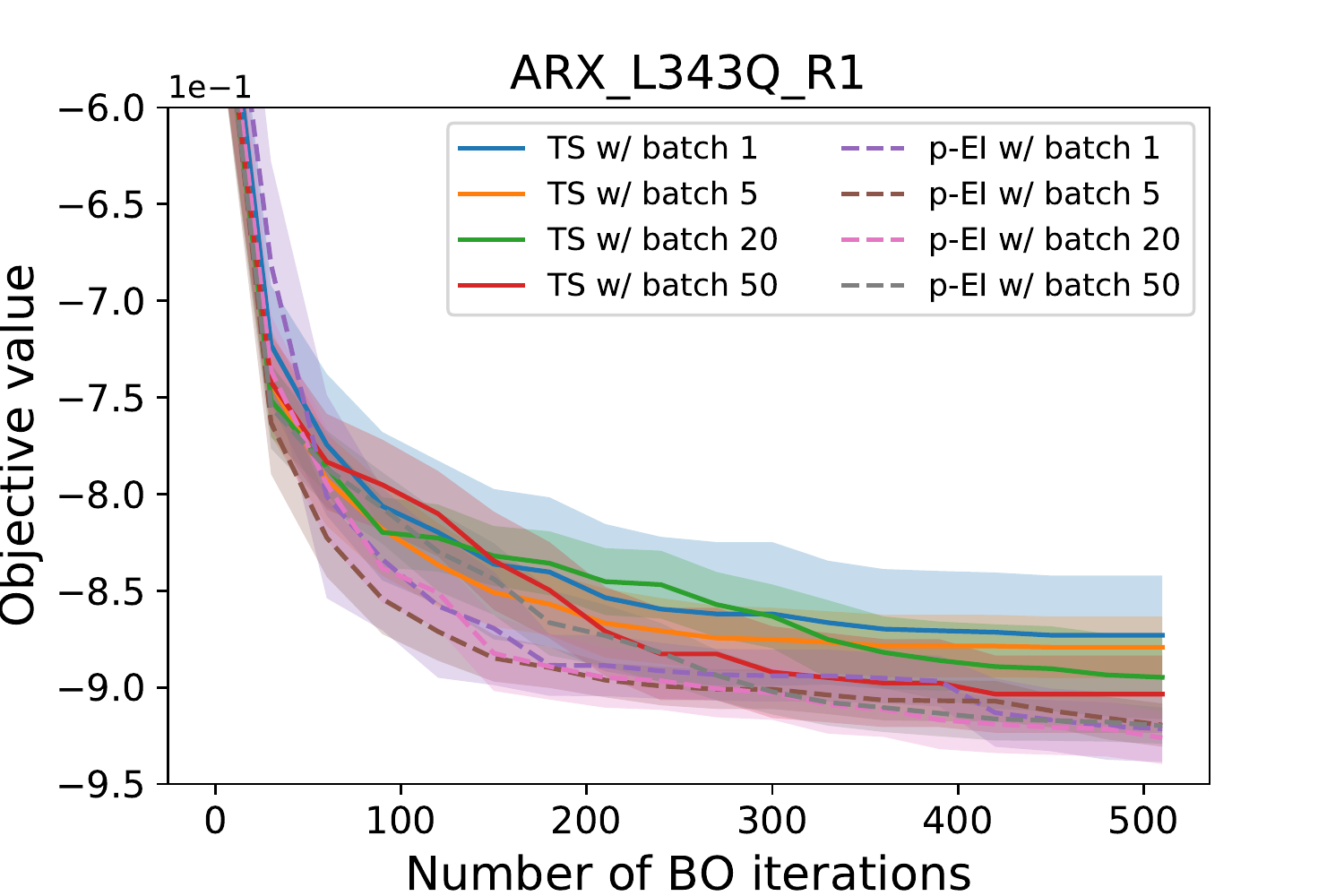}
\label{fig:bio_seq_overall}}
\caption{Representative results on biological sequence design problem for one transcription factor.} 
\label{fig:ts_versus_ei}
\end{figure*}

\begin{figure*}[h!]
\centering
\subfloat[Subfigure 2 list of figures text][]{
\includegraphics[width=0.33\textwidth]{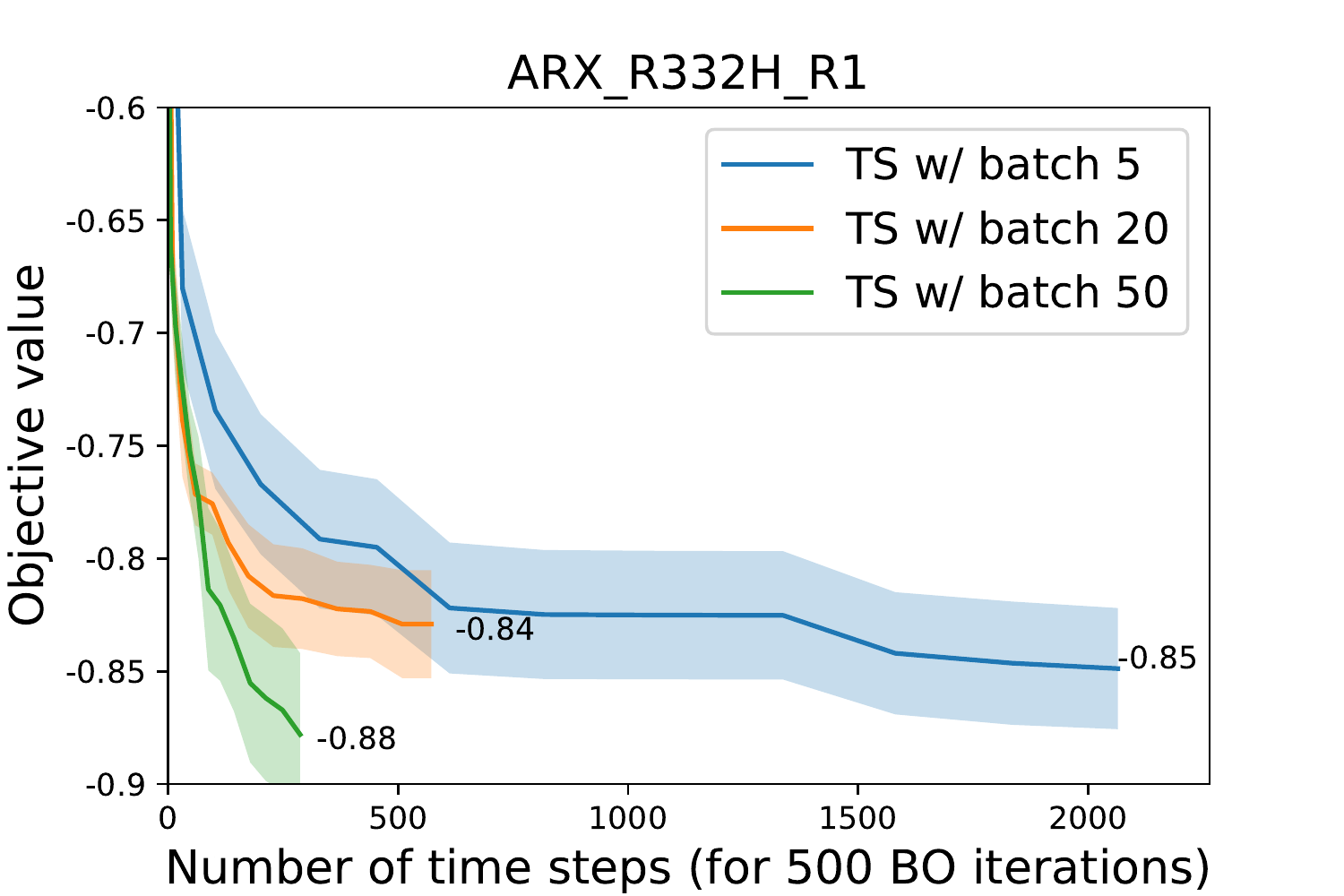}
\label{fig:ps1}}
\subfloat[Subfigure 1 list of figures text][]{
\includegraphics[width=0.33\textwidth]{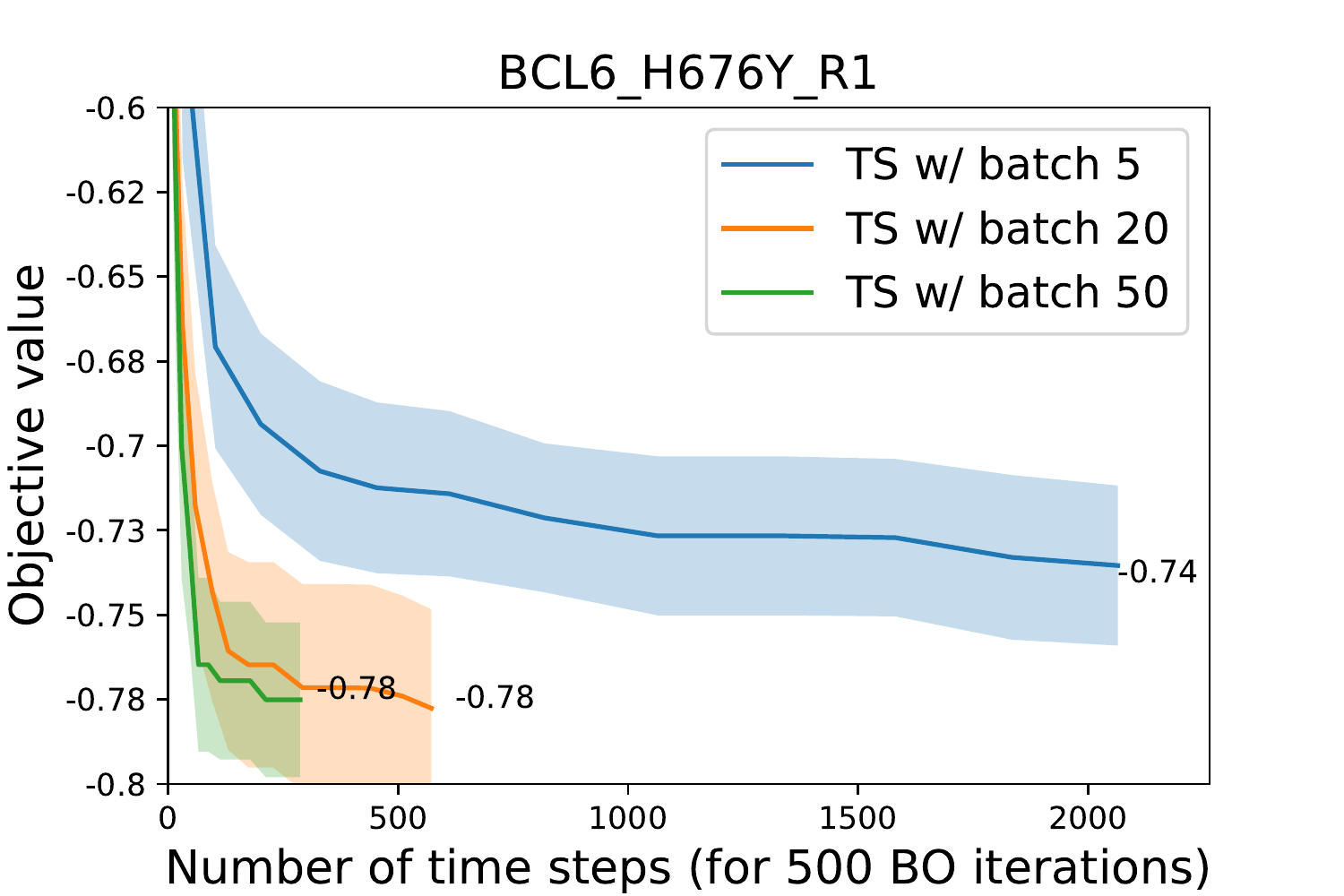}
\label{fig:ps2}}
\subfloat[Subfigure 1 list of figures text][]{
\includegraphics[width=0.33\textwidth]{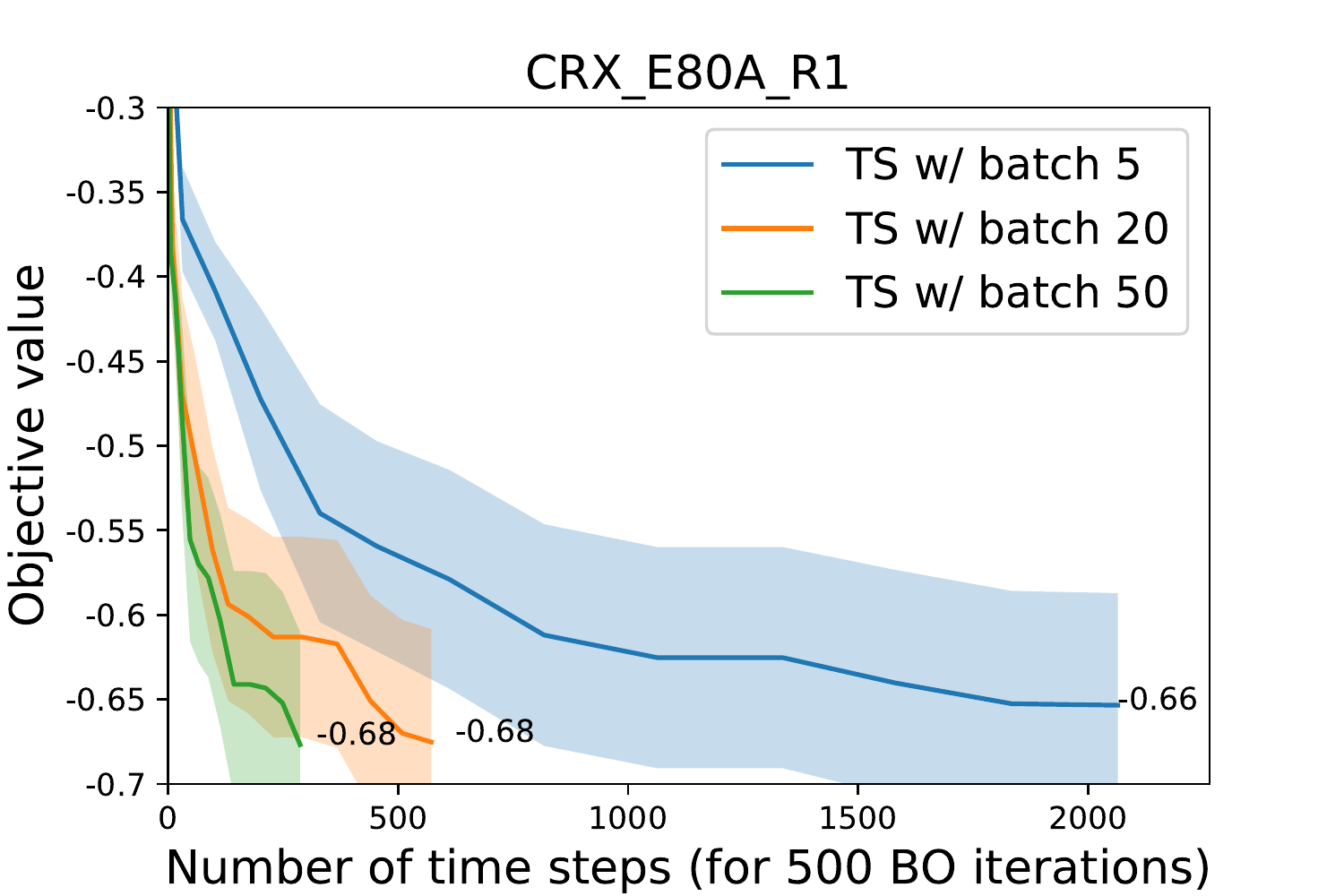}
\label{fig:ps3}}\\
\subfloat[Subfigure 2 list of figures text][]{
\includegraphics[width=0.33\textwidth]{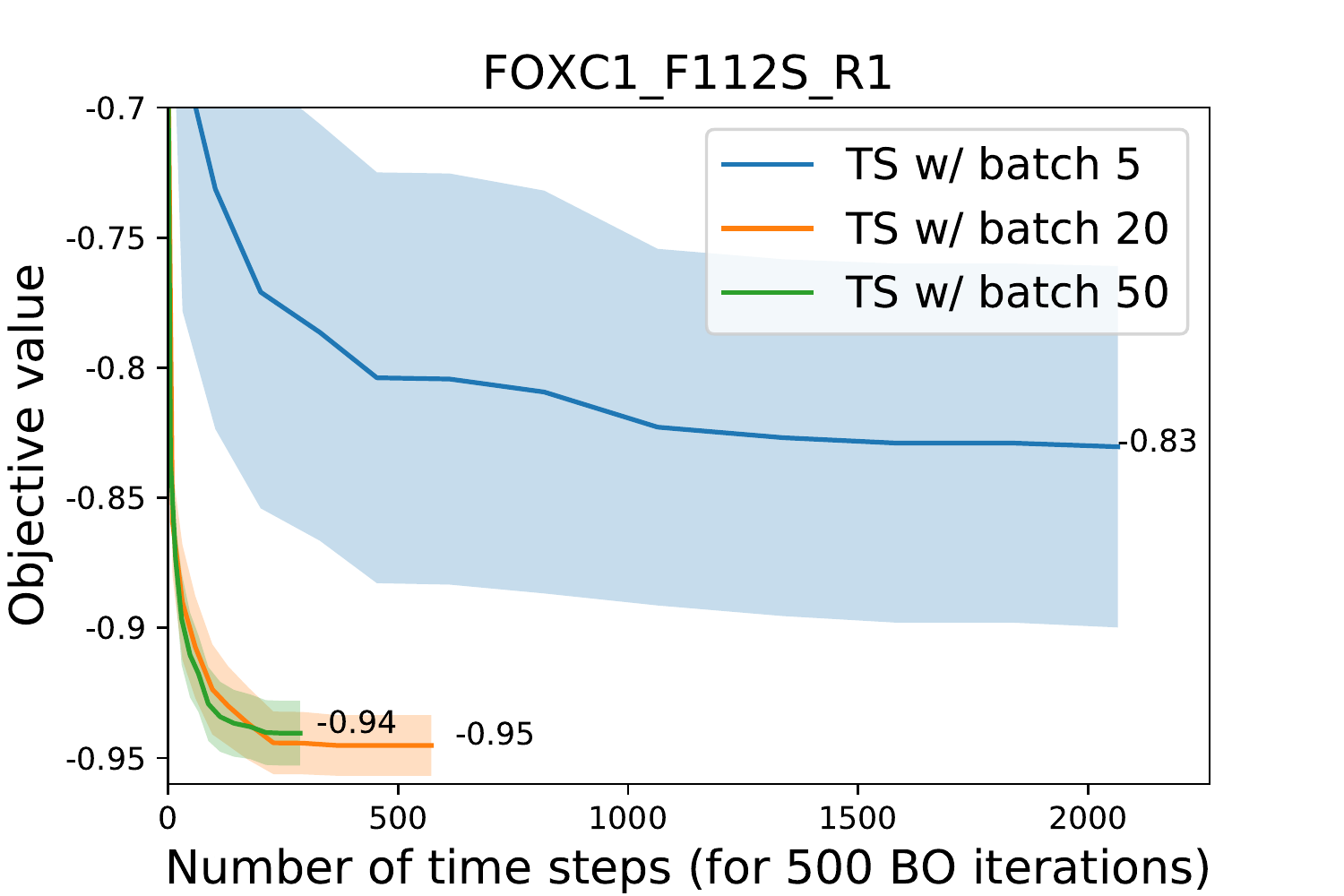}
\label{fig:ps4}}
\subfloat[Subfigure 1 list of figures text][]{
\includegraphics[width=0.33\textwidth]{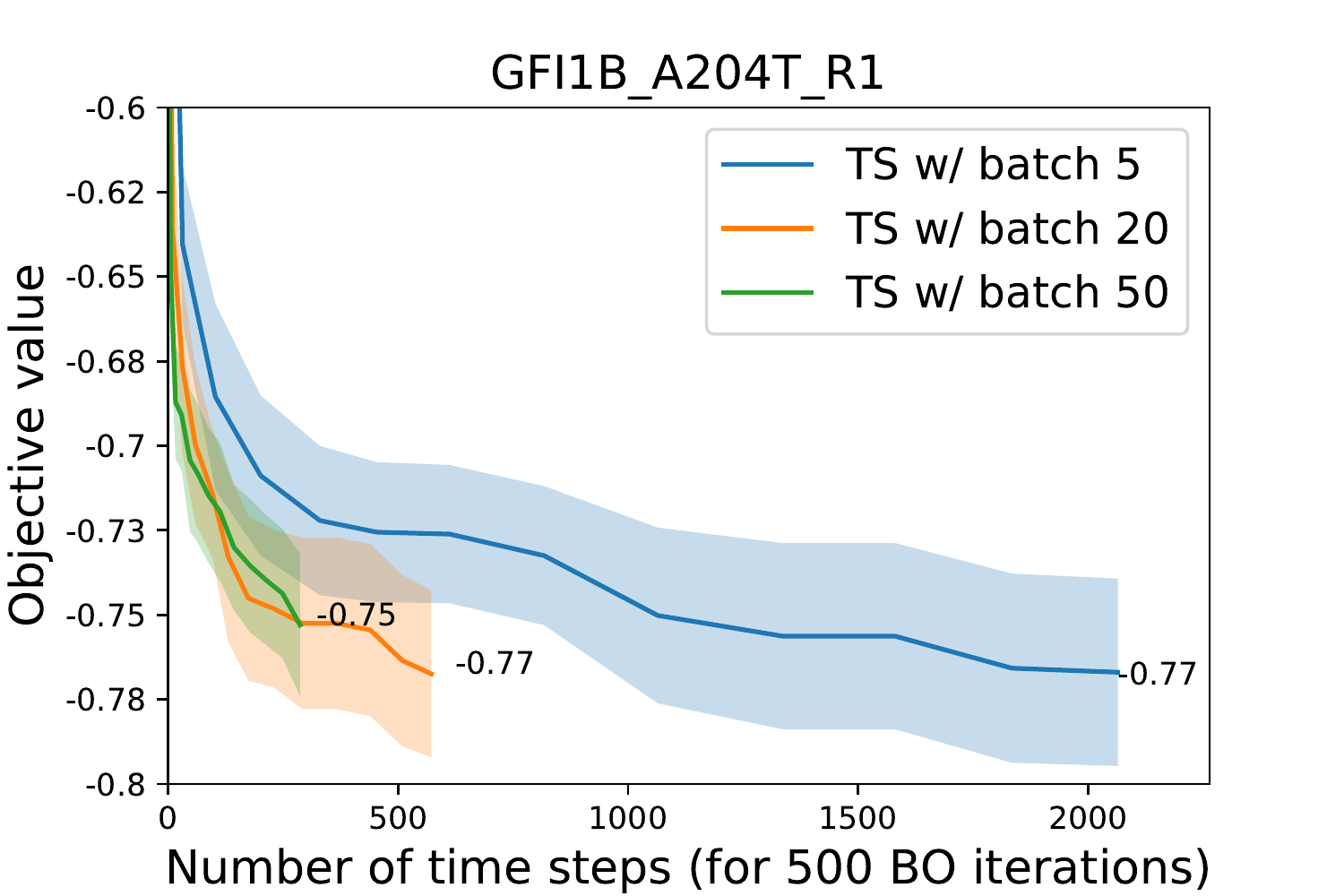}
\label{fig:ps5}}
\subfloat[Subfigure 1 list of figures text][]{
\includegraphics[width=0.33\textwidth]{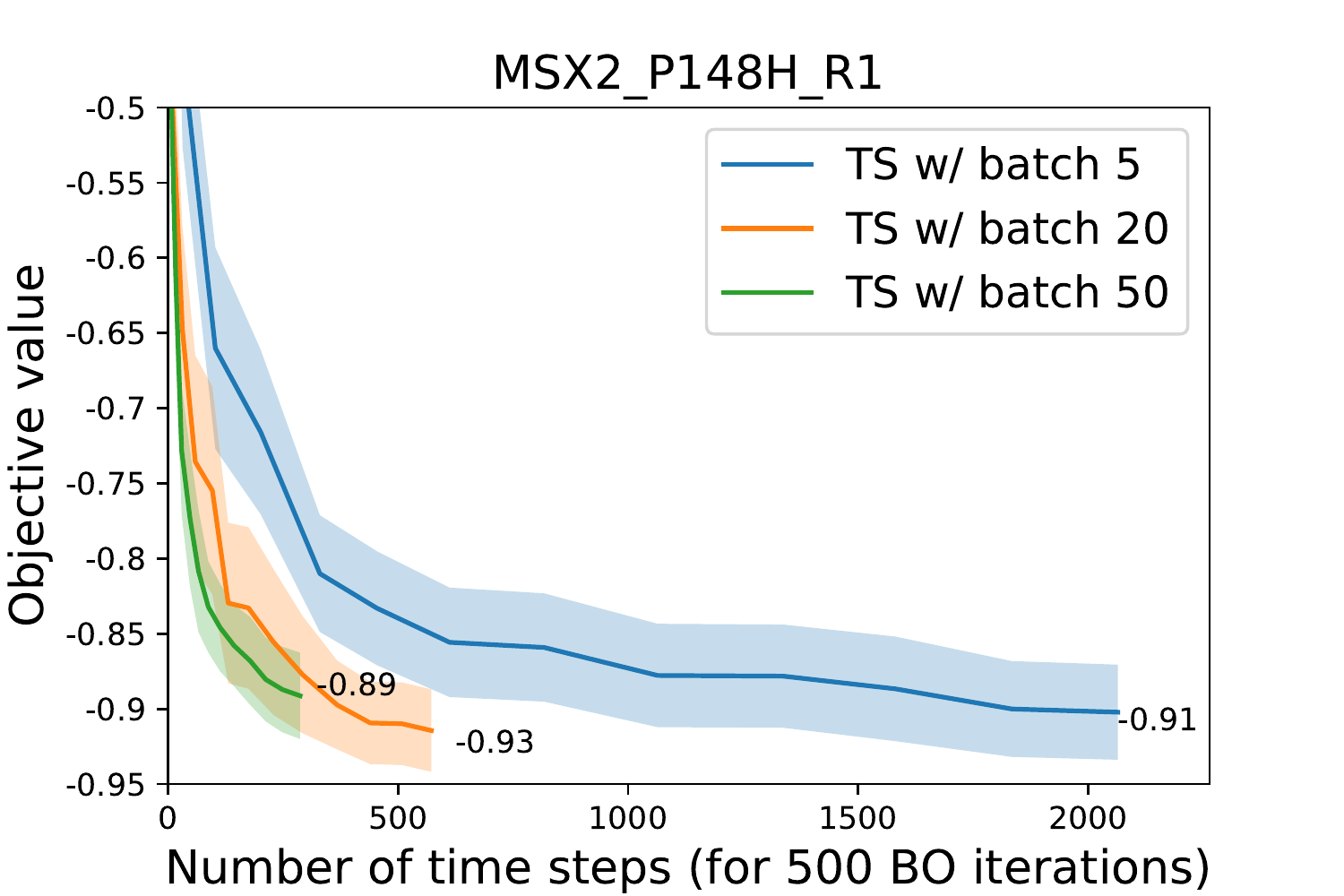}
\label{fig:ps6}}
\caption{Results on biological sequence design with Thompson sampling for six transcription factors.} 
\label{fig:ps}
\end{figure*}

\vspace{0.8ex}

\noindent {\bf Electrified aviation power system design.}  
We consider %optimizing 
the design of %electrified aviation 
power system for unmanned aerial vehicles (UAVs) via a time-based static simulator \cite{uav_design,belakaria2020PSD}. %The UAV system architecture consists of a central Li-ion battery pack, hex-bridge DC-AC inverters, PMSM motors, and necessary wiring. %(\cite{belakaria2020machine}).
Each structure is specified using five design variables such as the battery pack configuration %(battery cells in series or parallel) 
and motor
size. Evaluation of each design requires performing a computationally-expensive simulation. We consider two variants of this design task: % from a real-world perspective: 
 (1) Optimize {\em total energy}; and (2) Optimize {\em mass}. 
We employed a dataset of 250,000 candidate designs for our experiments. The categorical variables are encoded as $16$ bit binary variables.

To make the benchmark challenging, we initialized the surrogate models by randomly selecting from worst (in terms of objective) 10\% structures.  Results for this benchmark are shown in Figure \ref{fig:uav}. Interestingly, both BOCS and SMAC shows good performance on both mass and total energy objectives of this benchmark. Since EI has the tendency to be relatively more exploitative, COMBO shows poor convergence but reaches the best value at the end of allocated budget. MerCBO converges much faster on both problems, but the performance plateaus out on the mass objective. We attribute this behavior to the naturally exploratory behavior of the Thompson sampling acquisition function.

\subsection{Parallel Biological Sequence Design}

\vspace{0.5ex}

\noindent {\bf Motivation.} Design of optimized biological sequences such as DNA and proteins is a fundamental problem with many important applications \cite{yang_protein_review, angermueller2020population,yang_aistats}. These design problems have the following requirements: uncover a diverse set of structures ({\em diversity}); select a large batch of structures in each round to perform parallel evaluations ({\em large-scale parallel experiments}); and use parallel experimental resources to accelerate optimization ({\em real-time accelerated design}). 

\vspace{0.5ex}

\noindent {\bf Benefits of Thompson sampling (TS).} TS is a powerful approach that meets these requirements \cite{parallel_ts_lobato}. 
Any acquisition function is defined as the expectation of a utility function under the posterior predictive distribution $p(y|\mathbf{x}, D) = \int p(y|\theta) p(\theta|\mathbf{x}, D)$. TS approximates this posterior by a single sample $\theta^* \sim p(\theta|\mathbf{x}, D)$ which inherently enforces exploration due to the variance of this Monte-Carlo approximation. In MerCBO, we can sample as many $\theta$ (via Mercer features) %based approximation of the GP posterior) 
as required and can employ scalable AFO solvers for each sample in parallel. However, parallel version of EI (acquisition function employed in COMBO) does not meet most of the requirements as it selects the batch of structures for evaluation sequentially.

\vspace{0.5ex}

\noindent {\bf Experimental setup.} We evaluate TS and EI with GP based models on diverse DNA sequence design problems. The goal is to find sequences that maximize the binding activity between a variety of human transcription factors and every possible length-8 DNA sequence \cite{tfbind_domain,angermueller2020population}. Categorical variables are encoded as $2$ bit binary variables. We multiply objective values by -1 to convert the problem into minimization for consistency. We compare the parallel version of EI as proposed in \cite{BO:NIPS2012} and used in multiple earlier works \cite{parallel_ts_lobato,kandasamy2017asynchronous} with parallel TS.
For a batch of $B$ evaluations in each iteration, parallel-EI works by picking the first input in the same way as in the sequential setting and selects the remaining inputs $j$ = 2 to $B$ by maximizing the expected EI acquisition function under all possible outcomes of the first $j-1$ pending evaluations (aka %also called as 
fantasies \cite{BO:NIPS2012}).  %This is inherently a sequential procedure. 
On the other hand, TS is easily scalable and parallelizable by sampling $B$ $\theta$'s from the GP posterior and optimizing each of them independently with our MerCBO approach. We run both TS and EI experiments on a 32 core Intel(R) Core(TM) i9-7960X CPU @ 2.80GHz machine. All reported time units are in seconds on the same machine.

\vspace{0.5ex}

\noindent {\bf Discussion of results.} Figure \ref{fig:ts_versus_ei} shows the canonical comparison of parallel TS with parallel EI (p-EI) on one transcription factor from the DNA-binding affinity benchmark. The numbers within the plots show the mean objective value %found on each benchmark 
for a %given 
budget of 500 evaluations. Although parallel-EI is slightly better in terms of optimization performance, %(objective value of best uncovered design), 
TS is extremely fast and is useful for practitioners in time-constrained applications including drug and vaccine design. %As mentioned earlier, 
The diversity of best batch of sequences is equally important %in these domains 
to hedge against the possibility of some chosen candidate designs failing on downstream objectives \cite{angermueller2020population}. Figure \ref{fig:bio_seq_diversity} shows the results comparing the diversity of sequences (on mean Hamming distance metric) found by TS versus parallel-EI. There are two key observations that can be made from this figure. First, increasing the batch size increases the diversity of sequences. Second, TS finds comparatively more diverse sequences than parallel-EI. 

Interestingly, performance of parallel TS improves with increasing batch size. To justify this observation, we further evaluated parallel TS on six other transcription factors as shown in Figure \ref{fig:ps}. As evident from the figure, performance of parallel TS improves with increasing batch size on five out of six benchmarks. This shows that the exploratory behavior of TS, which can be bad in some sequential settings, helps in better performance for the parallel setting.

\section{Summary and Future Work}
We introduced an efficient approach called MerCBO for optimizing expensive black-box functions over discrete spaces. MerCBO is based on computing Mercer features for diffusion kernels and fast submodular relaxation based acquisition function optimization. We showed that MerCBO produces similar or better performance than state-of-the-art methods on multiple real-world problems. Future work includes using the proposed Mercer features for entropy based acquisition functions such as max-value entropy search.

\vspace{1.0ex}

%\footnotesize
\noindent {\bf Acknowledgements.} This research is supported in part by National Science Foundation (NSF) grants IIS-1845922, OAC-1910213, and CNS-1955353. The  views  expressed  are  those  of  the  authors and  do  not  reflect  the official policy or position of the NSF. Jana Doppa would like to thank Tom Dietterich for useful discussions.

\bibliography{references}
\newpage
\appendix
\begin{figure}[hbt!]
\centering
\subfloat[Subfigure 2 list of figures text][]{
\includegraphics[width=0.40\textwidth]{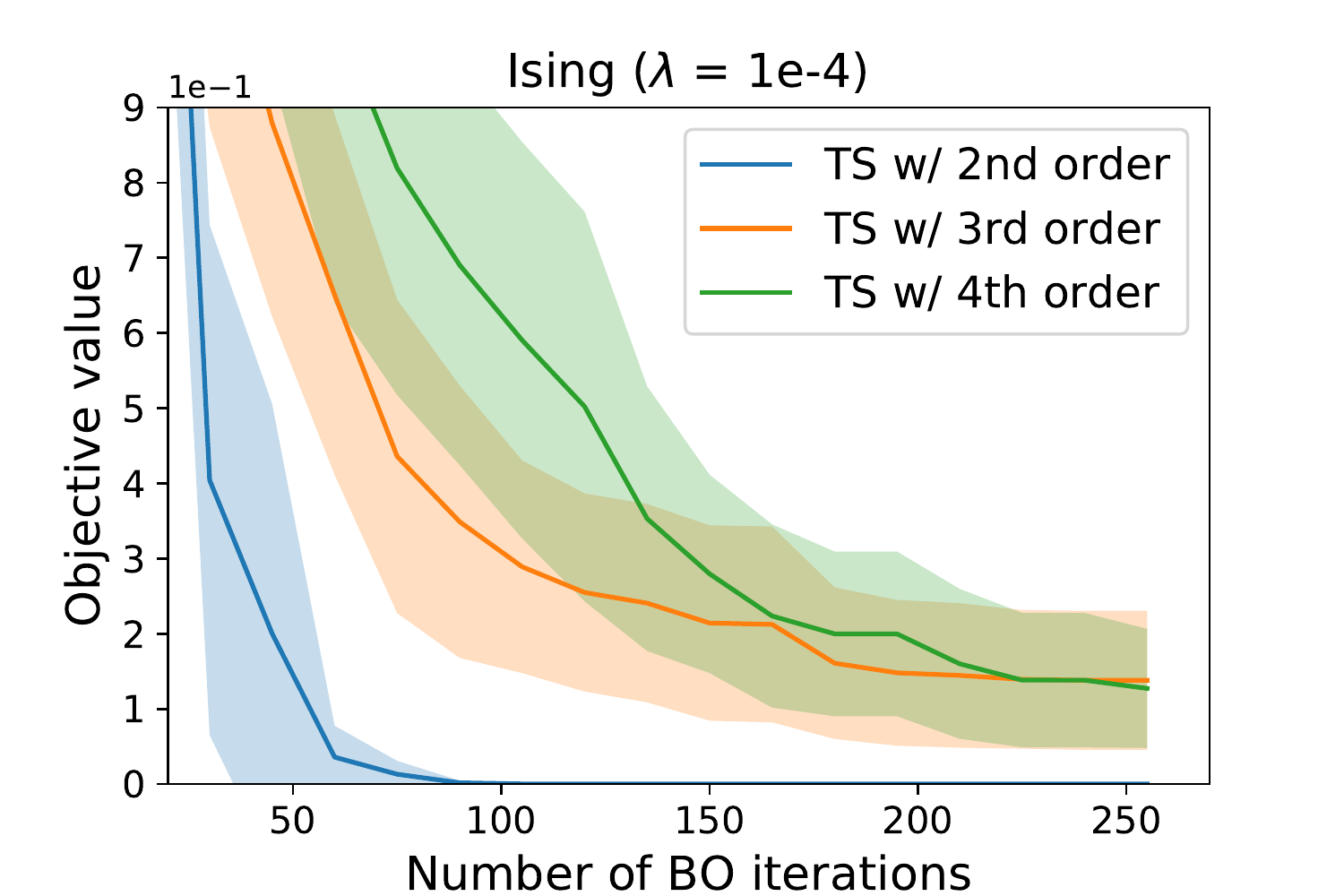}
\label{fig:ising_0_oc}}
\quad
\subfloat[Subfigure 1 list of figures text][]{
\includegraphics[width=0.40\textwidth]{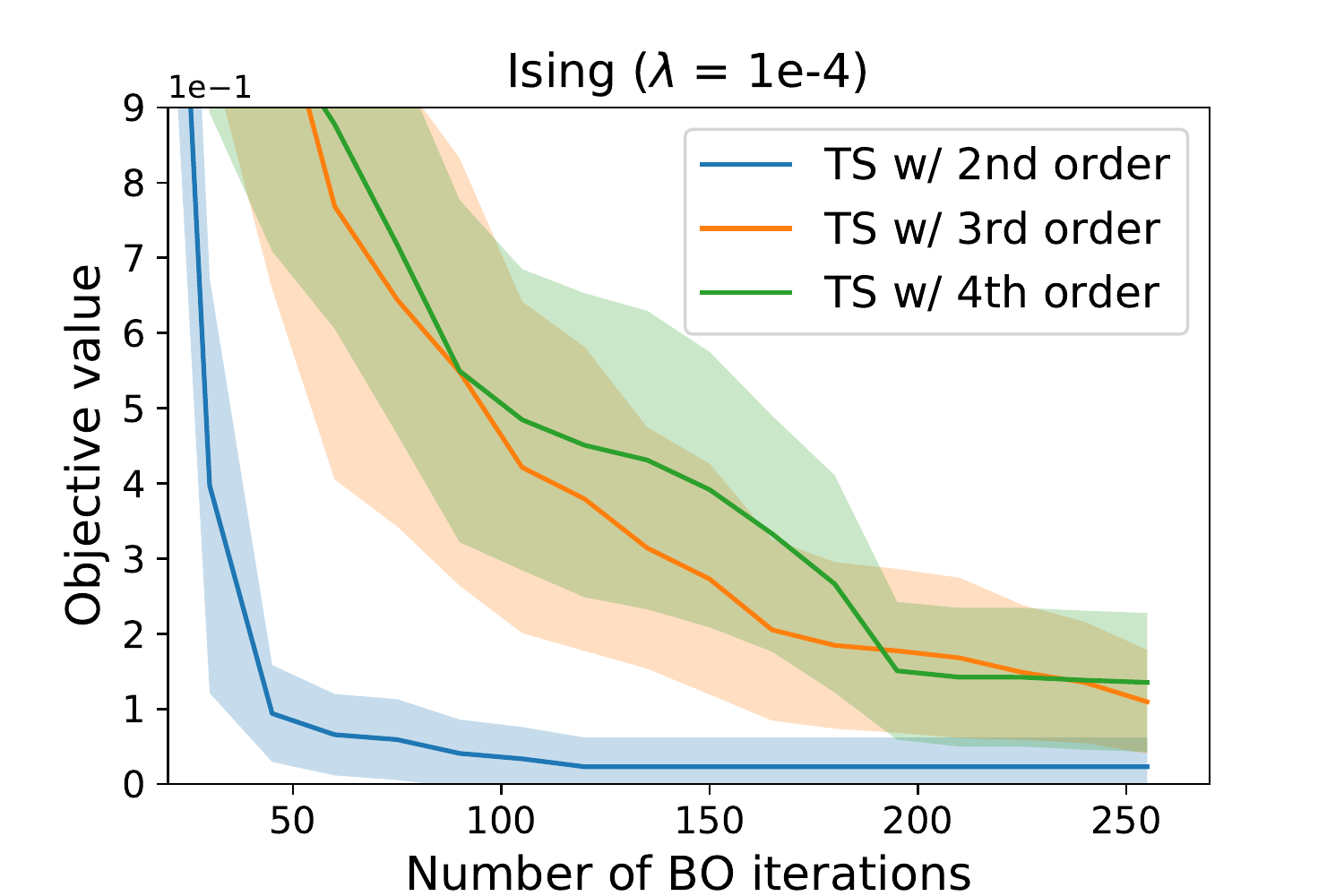}
\label{fig:ising_1e_4_oc}}
\quad
\subfloat[Subfigure 1 list of figures text][]{
\includegraphics[width=0.40\textwidth]{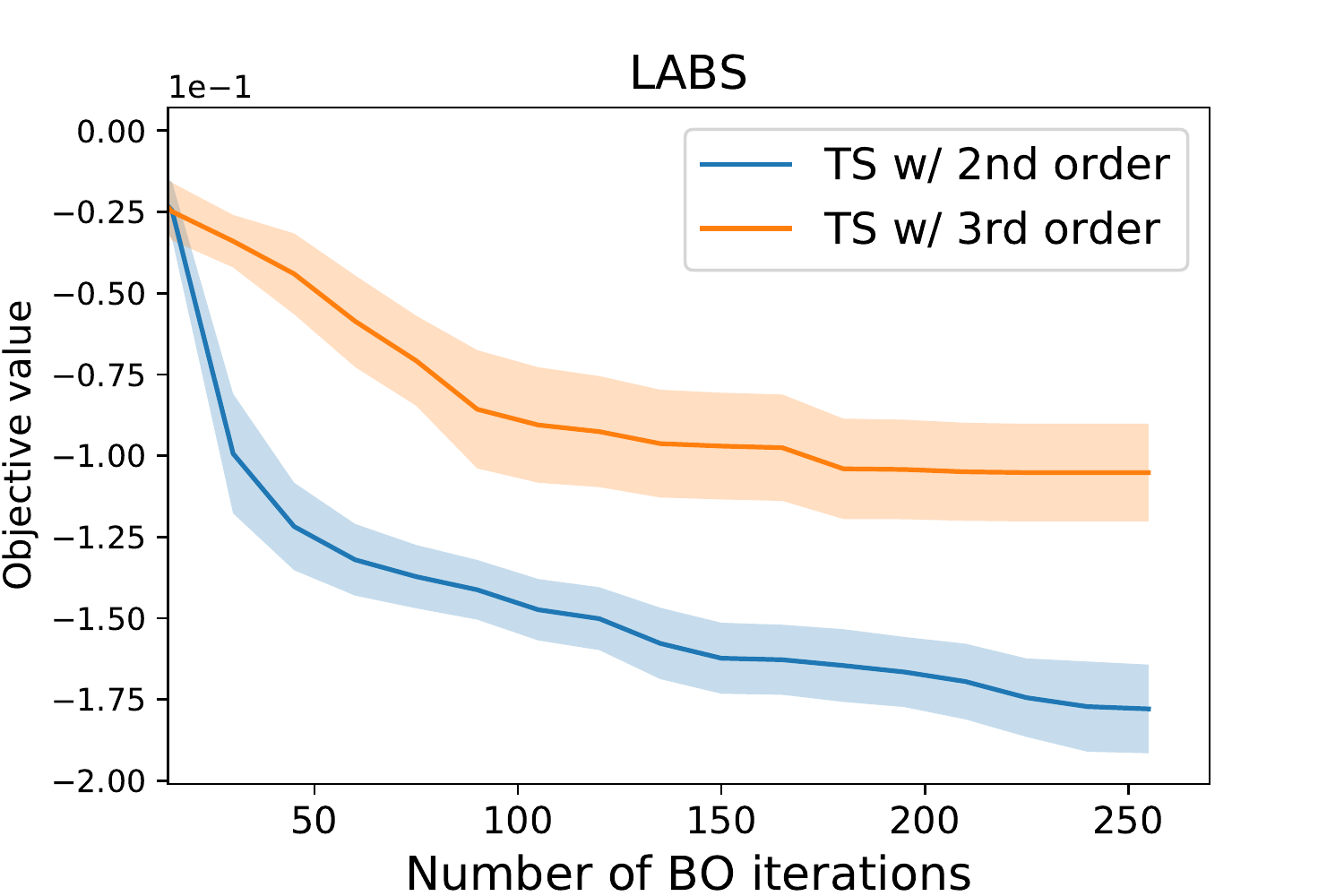}
\label{fig:labs_oc}}
% \subfloat[Subfigure 1 list of figures text][]{
% \includegraphics[width=0.35\textwidth]{figures/ising_1e_2.pdf}
% \label{fig:ising_1e_4}}
\caption{Results comparing overall BO performance using MerCBO with different orders of Mercer Features.} 
\label{fig:ising_oc}
\end{figure}

\begin{figure}[hbt!]
\centering
\subfloat[Subfigure 2 list of figures text][]{
\includegraphics[width=0.40\textwidth]{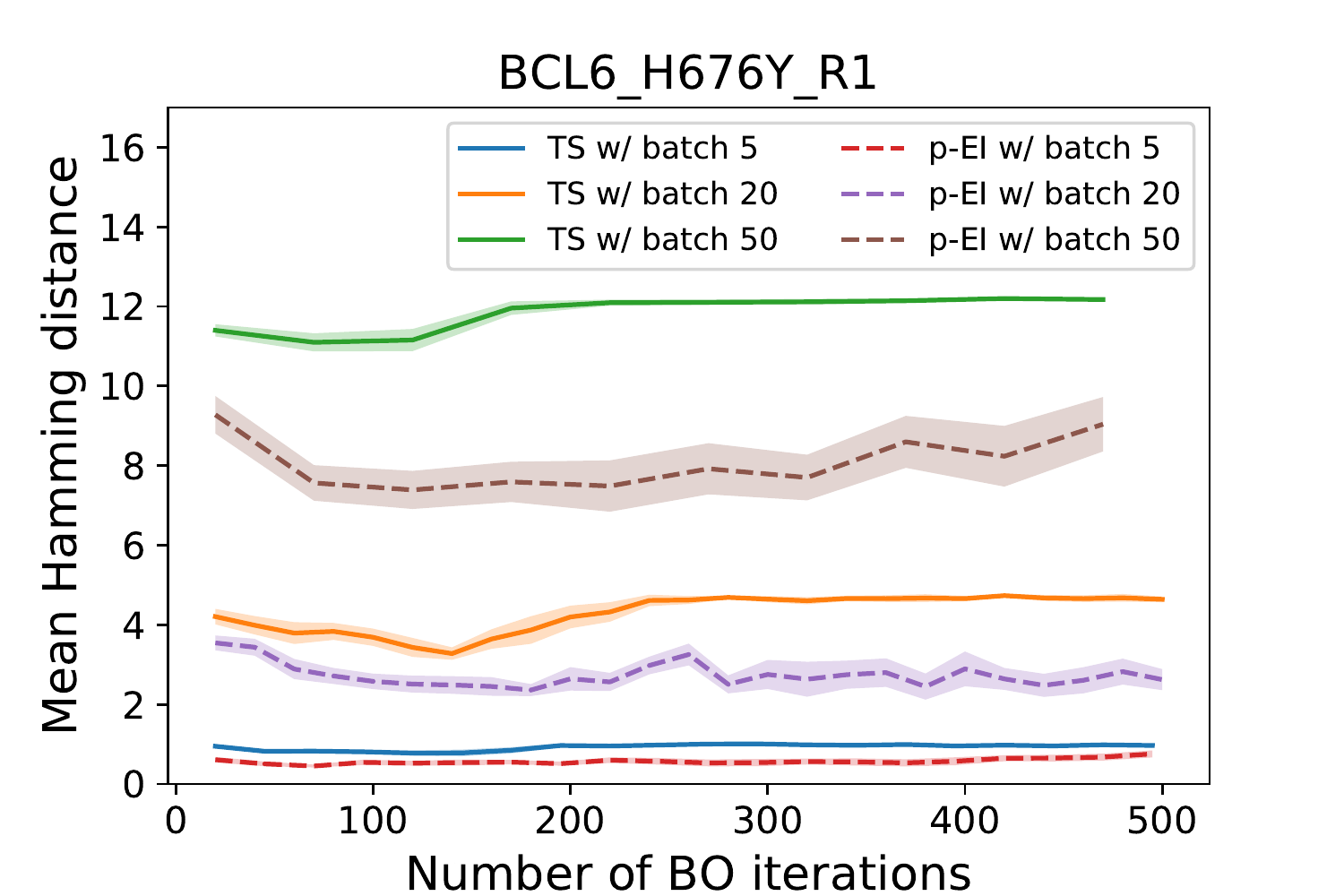}
\label{fig:bio_seq_diversity_1}}
\quad
\subfloat[Subfigure 1 list of figures text][]{
\includegraphics[width=0.40\textwidth]{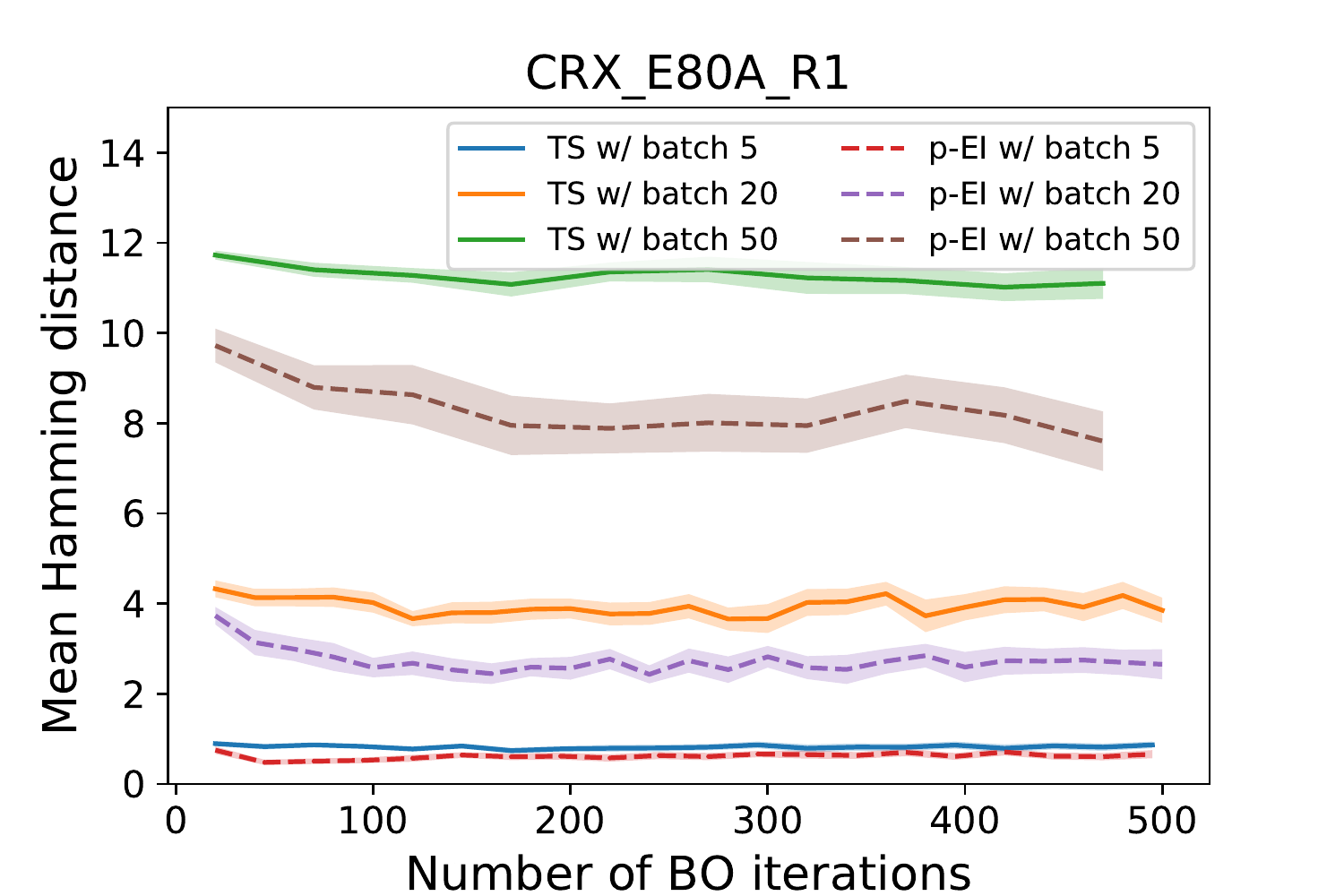}
\label{fig:bio_seq_diversity_2}}
\quad
\subfloat[Subfigure 2 list of figures text][]{
\includegraphics[width=0.40\textwidth]{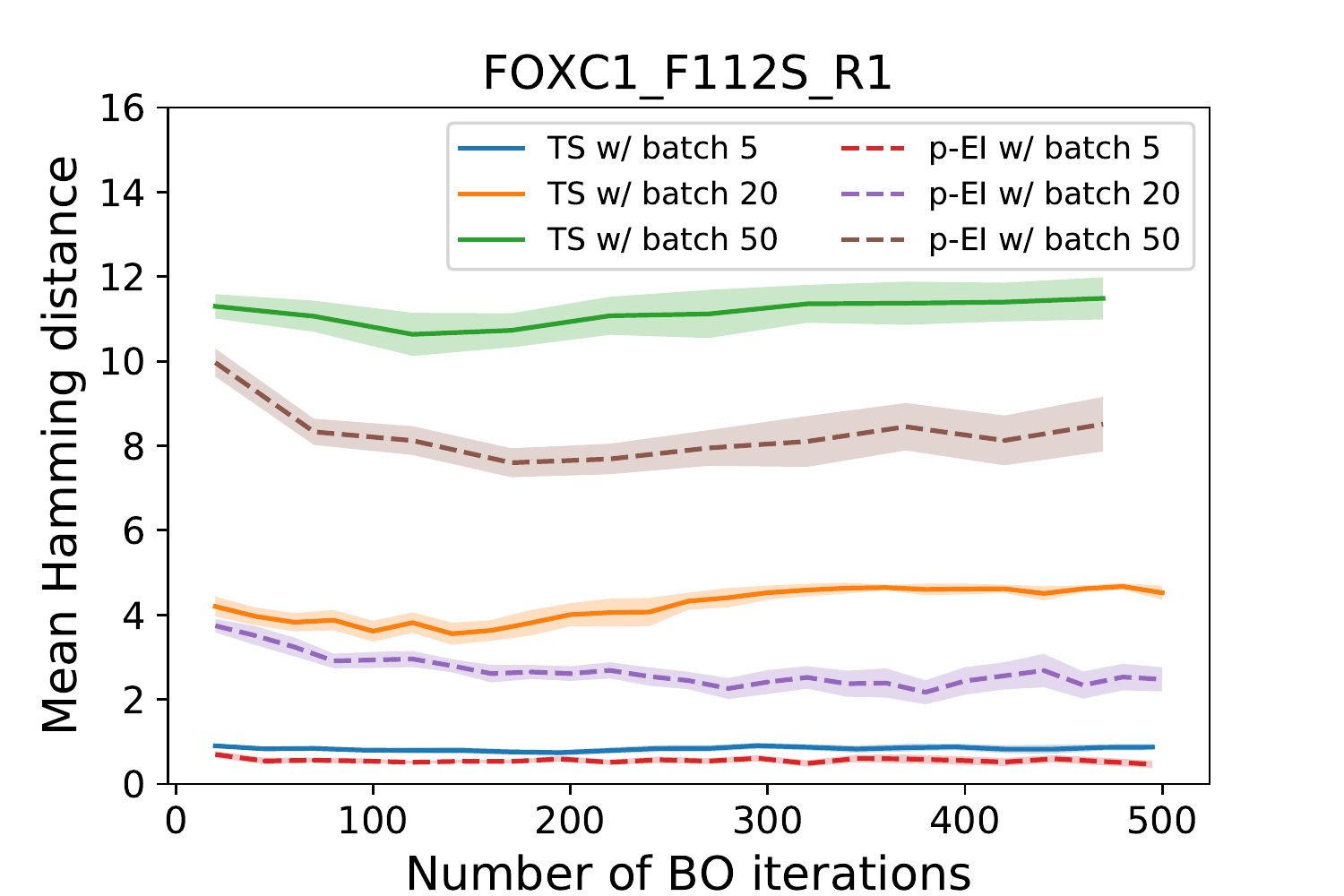}
\label{fig:bio_seq_diversity_3}}
\caption{Results comparing diversity of sequences generated by parallel TS vs. parallel EI on the biological sequence design problem for multiple transcription factors. Parallel TS generates significantly more diverse sequences with large batch sizes.} 
\label{fig:bio_seq_ts_ei}
\end{figure}

\section{Appendix}

\subsection{Additional Experimental Results}

Figure \ref{fig:ising_oc} compares the overall BO performance of MerCBO with different orders of Mercer features (\ref{full_feature_maps}). The acquisition function optimization for order greater than two is performed exactly same as COMBO \cite{COMBO}, i.e., local search with multiple random restarts.  We can clearly see from the figure that the performance degrades as order of Mercer features increases with $2$nd order giving the best results. These results further justifies our claim that $2$nd order Mercer features provide the best trade-off between tractability and performance.

Figure \ref{fig:bio_seq_ts_ei} shows more results comparing the sequences generated by parallel-TS versus parallel-EI on the {\em mean Hamming distance} metric to measure the diversity. As seen earlier in \ref{fig:ts_versus_ei}, the diversity of sequences increases with large batch size and TS generates significantly more diverse sequences than parallel-EI.

\end{document}